%% file: main.tex
\documentclass[lettersize,journal]{IEEEtran}
\usepackage{amsmath,amsfonts}
\usepackage{algorithmic}
\usepackage{array}
\usepackage[caption=false,font=normalsize,labelfont=sf,textfont=sf]{subfig}
\usepackage{textcomp}
\usepackage{stfloats}
\usepackage{url}
\usepackage{verbatim}
\usepackage{graphicx}
\def\BibTeX{{\rm B\kern-.05em{\sc i\kern-.025em b}\kern-.08em
    T\kern-.1667em\lower.7ex\hbox{E}\kern-.125emX}}
\usepackage{balance}

\usepackage{multirow}
\usepackage{graphicx}
\usepackage{amsmath}
\usepackage{amssymb}
\usepackage{mathtools, nccmath}
\usepackage{amsthm}
\usepackage{pifont}

\usepackage{algorithm, algorithmic}
\usepackage{graphicx}
\usepackage{wrapfig, lipsum, booktabs}
\usepackage{colortbl}
\usepackage{comment}
\usepackage{cleveref}
\usepackage{marvosym}
\usepackage{enumitem}
\usepackage{tabularx}
\usepackage{nicematrix}
\usepackage{caption}
\usepackage{diagbox}
\usepackage{dsfont}
\usepackage{natbib}

\begin{document}
\title{Curriculum Dataset Distillation}
\author{Zhiheng Ma\textsuperscript{*},
        Anjia Cao\textsuperscript{*},
        Funing Yang,
        Yihong Gong,~\IEEEmembership{Fellow,~IEEE,}
        Xing Wei\textsuperscript{\dag}
\IEEEcompsocitemizethanks{\IEEEcompsocthanksitem Z. Ma is with Shenzhen University of Advanced Technology, China, Guangdong Provincial Key Laboratory of Computility Microelectronics, China, and Shenzhen Institutes of Advanced Technology, Chinese Academy of Sciences, China (Email: mazhiheng@suat-sz.edu.cn). 
\IEEEcompsocthanksitem A. Cao, F. Yang, Y. Gong, and X. Wei are with School of Software Engineering, Xi'an Jiaotong University, China (Email: caoanjia7@stu.xjtu.edu.cn, moolink@stu.xjtu.edu.cn, ygong@mail.xjtu.edu.cn, and weixing@mail.xjtu.edu.cn).
\IEEEcompsocthanksitem $*$: Equal Contribution.%
\IEEEcompsocthanksitem $\dag$: Corresponding Author.}% <-this % stops a space
}

% \markboth{Journal of \LaTeX\ Class Files,~Vol.~18, No.~9, September~2020}%
% {How to Use the IEEEtran \LaTeX \ Templates}

\maketitle

\begin{abstract}
Most dataset distillation methods struggle to accommodate large-scale datasets due to their substantial computational and memory requirements. Recent research has begun to explore scalable disentanglement methods. However, there are still performance bottlenecks and room for optimization in this direction. In this paper, we present a curriculum-based dataset distillation framework aiming to harmonize performance and scalability. This framework strategically distills synthetic images, adhering to a curriculum that transitions from simple to complex. By incorporating curriculum evaluation, we address the issue of previous methods generating images that tend to be homogeneous and simplistic, doing so at a manageable computational cost. Furthermore, we introduce adversarial optimization towards synthetic images to further improve their representativeness and safeguard against their overfitting to the neural network involved in distilling. This enhances the generalization capability of the distilled images across various neural network architectures and also increases their robustness to noise. Extensive experiments demonstrate that our framework sets new benchmarks in large-scale dataset distillation, achieving substantial improvements of 11.1\% on Tiny-ImageNet, 9.0\% on ImageNet-1K, and 7.3\% on ImageNet-21K. Our distilled datasets and code are available at \url{https://github.com/MIV-XJTU/CUDD}.
\end{abstract}


\input{figures/baseline-ours}
\input{sections/intro}
\input{sections/method}

\input{sections/experiments}
\input{sections/related}
\input{sections/conclusion}
\input{sections/acknowledgements}

{
\small
\bibliographystyle{unsrt}
\bibliography{main}
}
\newpage
\input{sections/appendix}

\end{document}

%% file: figures/baseline-ours.tex
\begin{figure*}[!ht]
\centering
\includegraphics[width=\linewidth]{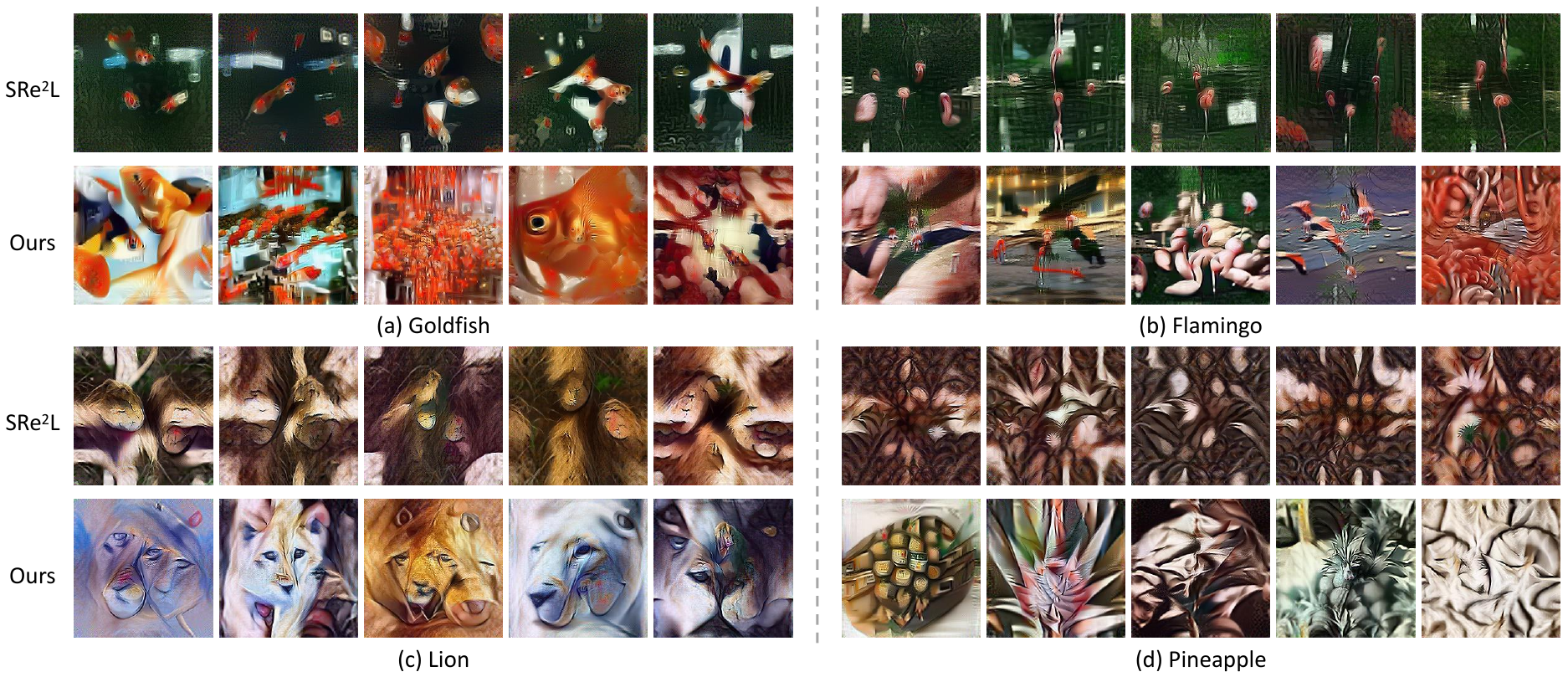}
% \vspace{4pt}
% \vspace{-15pt}
    \caption{\textbf{ImageNet-1K Distillation Comparison: our method vs. SRe$^2$L}~\cite{yin2023sre2l}. In contrast to SRe$^2$L, which often results in images with repetitive patterns, our approach creates synthetic images with a much richer diversity of patterns.}
    \label{fig:baseline-ours}
% \vspace{-.5em}
\end{figure*}

%% file: sections/intro.tex
\section{Introduction}
\label{sec:intro}
\IEEEPARstart{D}{ataset} distillation, as elucidated by \cite{wang2018dd}, entails compressing the original dataset into a significantly smaller synthetic dataset. This streamlined synthetic dataset confers pronounced advantages, particularly in improving data storage efficacy, fortifying privacy safeguards, and expediting training processes~\cite{zhao2021dc, cazenavette2022mtt, kim2022idc, wei2023speed, yin2023sre2l, guo2023datm}. Furthermore, the utility of this approach has been effectively demonstrated across various downstream applications, as evidenced in domains such as continual learning~\cite{wiewel2021condensed, deng2022rtp} and neural architecture search~\cite{such2020generative}.

A considerable number of algorithms for dataset distillation have approached the problem as a bi-level optimization task~\cite{wang2018dd,deng2022rtp}. This methodology involves an inner loop that focuses on updating the model and an outer loop for refining synthetic data. These strategies have shown significant advances in small-scale datasets such as MNIST and CIFAR~\cite{krizhevsky2009cifar}, which are known for their relatively low image resolutions and data volume. The outer loop, which evaluates the original data on the network trained with the synthetic data, ensures alignment between these two datasets. However, a major challenge arises from the high computational and memory costs associated with performing multiple unrolled iterations within the bi-level optimization framework. This limitation greatly hinders the application of bi-level-based methods to more complex real-world datasets like ImageNet~\cite{deng2009imagenet}.

Recently, Yin \emph{et al.} unveiled SRe$^{2}$L~\cite{yin2023sre2l}, an innovative approach that decouples bi-level optimization into discrete processes, achieving commendable dataset distillation efficacy on ImageNet. Initially, the method involves training a neural network on the original dataset, succeeded by applying the model inversion technique~\cite{mordvintsev2015deepdream, mahendran2015understanding, dosovitskiy2016inverting, yin2020deepinversion} to generate a synthetic dataset from the trained model. The last step involves the adoption of data augmentation and relabeling strategies~\cite{shen2022fkd} to substantially enhance dataset diversity. Remarkably, SRe$^{2}$L obviates the necessity for evaluation on the original dataset by leveraging the network, trained on the original data, as an effective surrogate. This strategy significantly decreases computational demands, thereby facilitating scalability to large-scale datasets.

However, our analysis highlights a significant concern related to the data diversity generated by SRe$^{2}$L, as depicted in Figure~\ref{fig:baseline-ours}. The recurrence of repetitive patterns in synthetic images leads to a decrease in the efficiency of SRe$^{2}$L, limiting its ability to comprehensively represent the original data distribution. Two key factors collectively contribute to this problem. First, SRe$^2$L lacks explicit guidance from evaluations conducted on the original dataset. Therefore, it cannot identify which parts of the samples should be further distilled. Second, the model inversion technique tends to generate patterns that are most representative -- or, in other terms, simpler -- from the perspective of the trained ``teacher" models. This leads to the synthetic dataset's insufficient exploration of complex patterns and the issue of image homogenization.

Building on this analysis, we propose a method called Curriculum Dataset Distillation (CUDD) that harmonizes scalability with representational diversity. This approach segments the creation of the synthetic dataset into a series of curricular stages, systematically synthesizing images in a progression from simple to complex to ensure comprehensive coverage of the original patterns. At the beginning of each curriculum, we first evaluate the original dataset through a ``student" neural network trained on all synthetic samples of prior curriculum to identify data instances that cannot be accurately classified, indicating areas where representational diversity is lacking.

However, the volume of the misclassified subset significantly exceeds the capacity of the synthetic dataset, necessitating further compression and representation by fewer synthetic samples. To maximize data efficiency in curriculum learning, we aim to increase the difficulty of the synthetic data as much as possible without compromising its alignment with the misclassified subset and its semantic correctness. The objective function is tripartite, encapsulating the essence of our method's innovation. The first component ensures that synthetic images are classified accurately by a ``teacher" network trained on the original dataset, thus ensuring the semantic correctness of the synthetic samples. The second component, an explicit regularization term, aligns the synthetic images with the intricacies of the misclassified subset, maintaining fidelity to the most challenging data aspects. Lastly, the adversarial loss from the ``student" network is strategically applied to further increase the difficulty and differentiate newly generated synthetic samples from all prior ones. As curriculum learning progresses, the capability of the ``student" network gradually approaches that of the ``teacher" network, providing increasingly effective adversarial feedback to generate more complex patterns.

Comprehensive experiments validate CUDD's superior performance, outstripping prior state-of-the-art methods across all real-world benchmark datasets. Specifically, CUDD achieves average improvements of 11.1\% on Tiny-ImageNet~\cite{le2015tiny}, 9.0\% on ImageNet-1K~\cite{deng2009imagenet}, and 7.3\% on ImageNet-21K~\cite{ridnik2021imagenet21k}, and notably doubles the performance on heterogeneous architectures such as DeiT-Tiny~\cite{touvron2021deit} and MLP-Mixer~\cite{tolstikhin2021mlpmixer}.

%% file: sections/method.tex
\section{Method}
\label{sec:method}
\subsection{Advantages and Limitations of Prior Methods}
Utilizing an extensive labeled dataset $\mathcal{T}=\left\{(x_i, y_i)\right\}_{i=1}^{\vert \mathcal{T} \vert}$, dataset distillation~\cite{wang2018dd, cui2022dcbench} aims to generate a significantly smaller synthetic dataset $\mathcal{S}=\left\{(\tilde{x}_i, y_i)\right\}_{i=1}^{\vert \mathcal{S} \vert}$, where $\vert \mathcal{S} \vert \ll \vert \mathcal{T} \vert$. In this context, $x$ denotes the original image, whereas $\tilde{x}$ signifies the synthetic image of identical resolution. The objective is to ensure that in the process of training any neural networks, the employment of synthetic datasets as substitutes for original datasets maintains equivalent performance while significantly diminishing the training costs. The dataset distillation can be formulated as a bi-level optimization task:
\begin{equation}
    \begin{aligned}
   \mathcal{S}^{*} &:= \mathop{\arg\min}\limits_{\mathcal{S}} \mathcal{L}_\text{ce}(\phi^{*}, \mathcal{T}) \\
   \text{s.t.} \quad \phi^{*} &:= \mathop{\arg\min}\limits_{\phi} \mathcal{L}_\text{ce}(\phi, \mathcal{S}).
    \end{aligned}
\end{equation}
In this framework, the inner optimization pertains to refining the proxy neural network $\phi$ using the synthetic dataset $\mathcal{S}$, whereas the outer optimization involves assessing the performance of this optimized network $\phi^{*}$ on the original dataset $\mathcal{T}$. However, achieving the optimal solution for this problem presents considerable challenges. This difficulty arises primarily because the inner optimization does not constitute a convex problem, and the sheer number of parameters in both the neural network and the synthetic dataset is substantial. Various methodologies, as referenced in the literature~\cite{wang2018dd,deng2022rtp}, employ implicit gradient techniques, calculating these gradients through back-propagation across the unrolled computational graph. Nonetheless, these methods frequently encounter significant computational and memory demands~\cite{zhao2021dc, zhao2023dm}, which pose substantial barriers when attempting to scale to extensive datasets like ImageNet~\cite{deng2009imagenet}.

Conversely, SRe$^2$L~\cite{yin2023sre2l} introduces a disentangled framework that directly generates the synthetic dataset via the model inversion technique~\cite{mordvintsev2015deepdream, mahendran2015understanding, dosovitskiy2016inverting, yin2020deepinversion}. This process is distinct in that it does not require feedback evaluation on the original dataset:
\begin{align}
   \theta^{*} &:= \mathop{\arg\min}\limits_{\theta} \mathcal{L}_\text{ce}(\theta, \mathcal{T}), \label{eq:squeeze}\\
   \mathcal{S}^{*}  &:= \mathop{\arg\min}\limits_{\mathcal{S}} \mathcal{L}_\text{ce+bn}(\theta^{*}, \mathcal{S}).\label{eq:recovory}
\end{align}
The above formulations succinctly elucidate the SRe$^2$L approach. Initially, in the phase described by Equation~\eqref{eq:squeeze}, the methodology begins with training a teacher network using the original dataset. Subsequently, the phase specified in Equation~\eqref{eq:recovory} is focused on generating the synthetic dataset from the teacher network. This generation process is accomplished by directly optimizing synthetic images from random noise. The overarching aim is to reconstruct the predictions and the batch statistics of the original dataset, employing both a standard cross-entropy loss $\mathcal{L}_\text{ce}$ and a batch statistic loss $\mathcal{L}_\text{bn}$:
\begin{equation}
    \mathcal{L}_{\text{bn}} = \sum\limits_{l} \Vert \mu_l(X_{\mathcal{S}}) - \text{BN}_l^{\mu} \Vert_2 + \Vert \sigma_l(X_{\mathcal{S}}) - \text{BN}_l^{\sigma} \Vert_2 ,
\end{equation}
where $\mu_l(X_{\mathcal{S}})$ and $\sigma_l(X_{\mathcal{S}})$ denote the batch statistics of the synthetic images preceding each batch normalization layer, with $l$ indicating the layer index. Furthermore, $\text{BN}_l^{\mu}$ and $\text{BN}_l^{\sigma}$ represent the respective trained parameters for each batch normalization layer.

Initially developed for white-box inverse attacks on neural networks~\cite{fredrikson2015model, he2019model}, the model inversion technique has since found widespread application in various fields, such as knowledge distillation~\cite{haroush2020knowledge, yin2020deepinversion, fang2021cmi}, network compression~\cite{cai2020zeroq, zhang2021dsg, zhong2022intraq}, and continual learning~\cite{smith2021always, liu2022few}, noted for its ability to replace the need for the original dataset. However, its application to dataset distillation encounters specific challenges: first, the absence of feedback evaluation on the original dataset implies that the synthetic dataset might not comprehensively represent the entire original data distribution; second, while model inversion is adept at producing highly representative patterns, it often falls short in exploring more complex and rare patterns. This limitation is evident in the comparison presented in Figure~\ref{fig:baseline-ours}. 
Building on this analysis, we propose a curriculum dataset distillation (CUDD) that integrates the strengths of previous methods. This strategy is founded on key principles: 1) reintroducing feedback evaluation of the original dataset while maintaining a controlled increment in computational expense, and 2) motivating the model to generate a diverse array of samples, facilitating exploration from simpler to more complex patterns.

\subsection{Curriculum Dataset Distillation}
\input{figures/overview}
We commence by segregating the synthetic dataset into $J$ distinct, non-overlapping subsets, expressed as $\bigcup_{j=1}^{J} \mathcal{S}_j = \mathcal{S}$, with $J$ signifying the aggregate number of curricula. The strategy is structured to sequentially generate synthetic subsets, progressing through each curriculum in turn. 

\textbf{Curriculum Feedback Evaluation}. Initially, a teacher neural network is trained, with its corresponding weights denoted as $\theta^{*}$, employing the complete original dataset $\mathcal{T}$, as explicated in Equation~\eqref{eq:squeeze}. In the bi-level optimization method, each iteration necessitates an evaluation of the original dataset, a process that presents significant scalability challenges. In contrast, SRe$^2$L omits explicit evaluation of the original dataset. We introduce a mediating strategy between these two methods, implementing feedback evaluation at the onset of each curriculum:
\begin{equation}
   \text{Init } \mathcal{S}_{j} \text{ with }  \mathcal{T}_{j} \sim R(\theta^{*}, \mathcal{T} \setminus \mathcal{T}_{1:j-1}) \cap W(\phi_{j-1}^{*}, \mathcal{T} \setminus \mathcal{T}_{1:j-1}),
   \label{eq:evaluate}
\end{equation}
where $\phi_{j-1}^{*}$ symbolizes the weight of the preceding student neural network trained on $\mathcal{S}_{1:j-1}^{*}$. The function $R(\theta^{*}, \mathcal{T} \setminus \mathcal{T}_{1:j-1})$ selects correctly classified samples from $\mathcal{T} \setminus \mathcal{T}_{1:j-1}$ using the teacher network $\theta^{*}$. Conversely, $W(\phi_{j-1}^{*}, \mathcal{T} \setminus \mathcal{T}_{1:j-1})$ identifies misclassified samples, acting as the erroneous subset selector.

As delineated in Equation~\eqref{eq:evaluate}, we perform a random sampling without replacement of a subset $\mathcal{T}_{j}$ from instances that are correctly classified by the teacher model yet misclassified by the preceding student model.  Importantly, the subset $\mathcal{T}_{j}$ is meticulously curated to mirror the size of $\mathcal{S}_{j}$, thereby ensuring that $\vert\mathcal{T}_{j}\vert=\vert\mathcal{S}_{j}\vert$. This facilitates the initialization of $\mathcal{S}_{j}$ with the elements of $\mathcal{T}_{j}$. 
This strategy, grounded in the easy-to-hard training paradigm common in curriculum learning~\cite{bengio2009curriculumlearning, kumar2010self, wang2024brain, ding2024towards}, presents multiple advantages: it ensures synthetic image diversity by progressively increasing difficulty, facilitates foundational feature learning with simpler images initially using a limited set, addresses more complex situations with a larger image pool in later stages, and permits reusing previously generated synthetic images, eliminating the need for their regeneration.

\input{algorithms/algorithm}
\textbf{Adversarial Data Optimization}. 
Typically, given the significantly lower quantity of synthetic images compared to the original images, the subset \(\mathcal{T}_{j}\) can cover only a minimal fraction of the misclassified images. Solely employing the previously mentioned initialization method is markedly insufficient. Consequently, we introduce our synthetic data optimization objective, which integrates an adversarial loss to further increase the difficulty and data efficiency of each synthetic image:
\begin{equation}
\begin{aligned}
     \mathcal{S}^{*}_{j} :&= \mathop{\arg\min}\limits_{\mathcal{S}_{j}} \mathcal{L}_\text{ce+bn}(\theta^{*}, \mathcal{S}_{j}) + \alpha_\text{reg} \mathcal{L}_\text{reg}(\mathcal{T}_{j}, \mathcal{S}_{j})\\ &+ \alpha_\text{adv}  \mathcal{L}_\text{adv}(\phi_{j-1}^{*}, R(\theta^{*}, \mathcal{S}_{j})). 
    \end{aligned}
    \label{eq:main}
\end{equation}
Our objective encompasses three components; the first component aligns with SRe$^2$L (Equation~\eqref{eq:recovory}), focusing on distilling synthetic images through model inversion from the teacher network, which is trained on the original dataset. From the curriculum learning viewpoint, this goal further suggests that the synthetic images are expected to be accurately classified by the teacher network. The second component is a regularization loss, ensuring that $\mathcal{S}_{j}$ remains aligned with its initial state $\mathcal{T}_{j}$. This is realized through the adoption of a Mean Squared Error (MSE) loss, applicable within either the pixel or the teacher network's feature space. Our experimental results reveal that the regularization loss $\mathcal{L}_\text{reg}$ significantly enhances diversity and acts as a complement to $\mathcal{L}_\text{ce+bn}$. While $\mathcal{L}_\text{ce+bn}$ ensures that $\mathcal{S}$ is derived from a proxy distribution akin to the original data distribution, it does not ensure the uniqueness of each sample. Conversely, $\mathcal{L}_\text{reg}$, by sampling $\mathcal{T}_{j}$ from $\mathcal{T}$ without replacement, ensures each synthetic sample's distinctiveness.

The third component of our approach is the adversarial loss. Instead of employing the cross-entropy loss, we utilize the non-saturating loss, which has been demonstrated to be more effective in adversarial training contexts, as evidenced in the literature~\cite{goodfellow2014gan, suzuki2022teachaugment}:
\begin{equation}
  \begin{aligned}
    \mathcal{L}_\text{adv} = -\frac{1}{N} \sum_{i=1}^{N} \sum_{c=1}^{C} \mathds{1}[y_i = c] \log(1-F(\phi_{j-1}^{*}, \tilde{x}_i)^{[c]}),& \quad\\ \text{s.t.}  \quad (\tilde{x}_i, y_i) \in R(\theta^{*}, \mathcal{S}_{j}),& 
  \end{aligned}
\end{equation}
where  $N$ signifies the batch size, $C$ stands for the overall class count, $(\tilde{x}_i, y_i)$ denotes the pair of a synthetic image and its associated label, and $F(\phi_{j-1}^{*}, \tilde{x}_i)$ signifies the softmax output of the student network. This objective is to further push the synthetic images towards the decision boundary to improve the informativeness of the synthetic images, as visualized in Figure~\ref{fig:tsne}. It should be emphasized that the adversarial loss is selectively applied to images accurately identified by the teacher network, denoted as $R(\theta^{*}, \mathcal{S}_{j})$. This approach ensures the prevention of synthetic image collapse.

\textbf{Student Network Training}. Following the generation of synthetic images for the current curriculum, we integrate these with all synthetic images produced in previous curricula to train the new student network for the subsequent curriculum. To further augment diversity, we also implement the relabeling technique~\cite{shen2022fkd} utilized in SRe$^2$L:
\begin{equation}
    \phi_{j}^{*} := \mathop{\arg\min}\limits_{\phi} \mathcal{L}_\text{ce}(\phi, \mathcal{S}^{*}_{1:j}).
\end{equation}
It is noteworthy that once the complete synthetic dataset is generated, both the teacher and student networks can be discarded, retaining solely the synthetic dataset for downstream tasks. The entirety of our algorithm is delineated in Algorithm~\ref{alg}. Given that the initial curriculum lacks a trained student network, we omit adversarial loss and select $\mathcal{T}_{1}$ from $\mathcal{T}$ only via the correct subset selection function.

%% file: figures/overview.tex
\begin{figure*}[t]
\centering
\includegraphics[width=\linewidth]{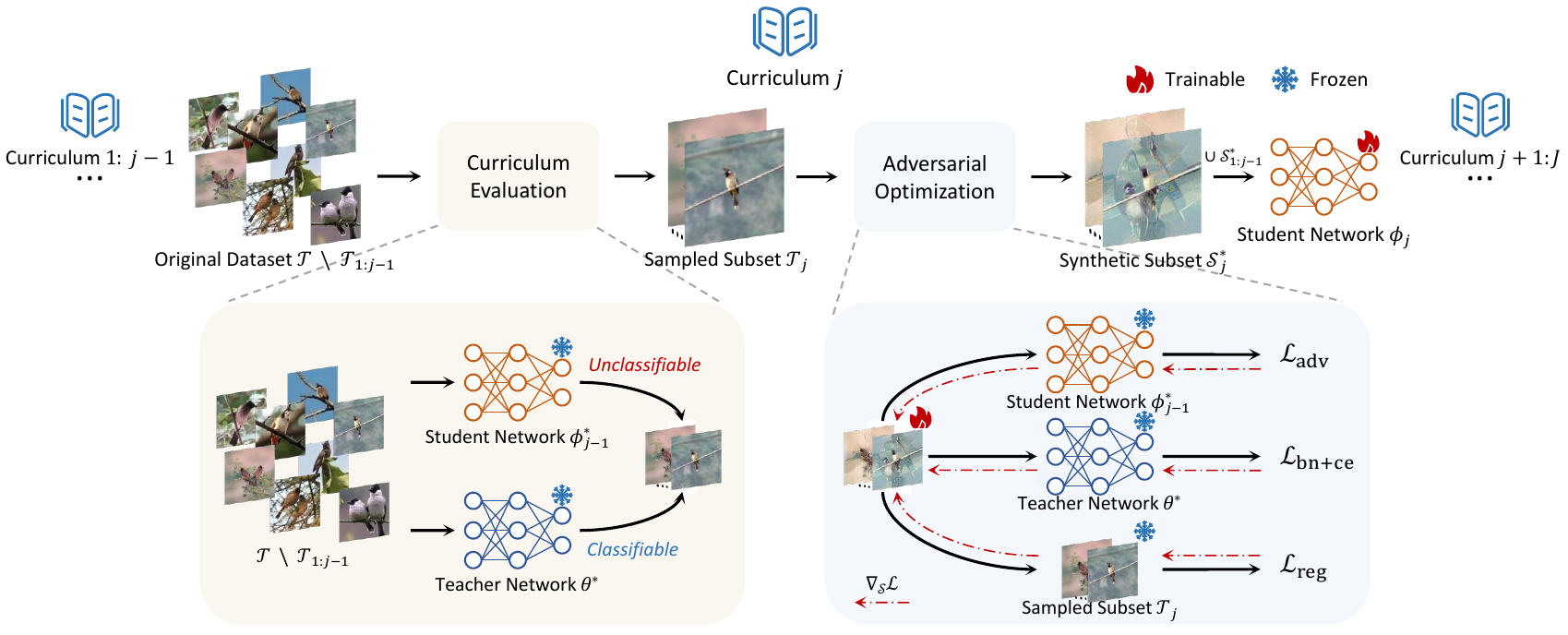}
% \vspace{4pt}
% \vspace{-12pt}
    \caption{\textbf{Curriculum Dataset Distillation Overview}. A single curriculum comprises three key phases: 1) Initial selection of samples from the original dataset that are misclassified by the previous curriculum's student network but correctly identified by the teacher network, serving as seeds for synthetic sample generation. 2) Optimization of synthetic samples using the objective function detailed in Equation~\eqref{eq:main}. 3) Integration of both existing and newly synthesized samples to train an updated student network.}
    \label{fig:overview}
% \vspace{-1em}
\end{figure*}

%% file: algorithms/algorithm.tex
\renewcommand{\algorithmiccomment}[1]{$\triangleright$ #1}
\begin{algorithm}[tb]
   \caption{Curriculum Dataset Distillation}
   \label{alg:cuda}
\begin{algorithmic}
   \STATE {\bfseries Input:} original dataset $\mathcal{T}$, pre-trained teacher network ${\theta}^{*}$, number of curricula $J$
   % \REPEAT
   \FOR{$j=1$ {\bfseries to} $J$}
   \IF{$j = 1$}
       \STATE \COMMENT{Omit the adversarial loss in Equation.~\eqref{eq:main}}
       \STATE \COMMENT{Initialize $\mathcal{S}_j$ via correct subset selection:}
        \STATE \quad $\mathcal{T}_{j} \sim R(\theta^{*}, \mathcal{T})$, $\mathcal{S}_{j} \leftarrow \mathcal{T}_j$
   \ELSE
        \STATE \COMMENT{Initialize $\mathcal{S}_j$ via feedback evaluation on $\phi_{j-1}^{*}$:}
        \STATE \quad $\mathcal{T}_{j} \sim R(\theta^{*}, \mathcal{T} \setminus \mathcal{T}_{1:j-1}) \cap W(\phi_{j-1}^{*}, \mathcal{T} \setminus \mathcal{T}_{1:j-1})$
        \STATE \quad $\mathcal{S}_{j} \leftarrow \mathcal{T}_j$
   \ENDIF
   \REPEAT
   \STATE \COMMENT{Compute distillation loss $\mathcal{L}(\theta^{*}, \phi_{j-1}^{*}, \mathcal{T}_j, \mathcal{S}_j)$ according to Equation.~\eqref{eq:main}}
   \STATE \COMMENT{Update $\mathcal{S}_{j}$ with respect to the loss:}
   \STATE \quad $\mathcal{S}_{j} \leftarrow \mathcal{S}_{j} - \nabla_{\mathcal{S}_{j}} \mathcal{L}(\theta^{*}, \phi_{j-1}^{*}, \mathcal{T}_j, \mathcal{S}_j)$
   \UNTIL{$\mathcal{S}_{j}$ converged, $\mathcal{S}_{j}^{*} \leftarrow \mathcal{S}_{j}$ }
   \STATE \COMMENT{Unify the synthetic dataset:} 
   \STATE \quad $\mathcal{S}_{1:j}^{*} \leftarrow \mathcal{S}_{1:j-1}^{*} \cup \mathcal{S}_{j}^{*}$
   \STATE \COMMENT{Train the student network on $\mathcal{S}_{1:j}^{*}$ to get ${\phi}_{j}^{*}$}
   \ENDFOR
   \STATE {\bfseries Output:} The final synthetic dataset $\mathcal{S} \leftarrow \mathcal{S}_{1:J}^{*} $
\end{algorithmic}\label{alg}
\end{algorithm}

%% file: sections/experiments.tex
\section{Experiments}
\label{sec:experiments}
\subsection{Dataset and Implementation Details}
% We evaluate CUDD on various datasets, including : Tiny-ImageNet~\cite{le2015tiny}, ImageNet-1K~\cite{deng2009imagenet}, ImageNet-21K-P~\cite{ridnik2021imagenet21k}, CIFAR-10 and CIFAR-100~\cite{krizhevsky2009cifar}.
We evaluate our methods on the following datasets: 
\begin{itemize}[topsep=1pt, partopsep=9pt, itemsep=-1pt, parsep=0.5ex]
    \item TinyImageNet~\cite{le2015tiny} is a 64$\times$64 dataset with 200 classes. Each class has 500 images for training and 50 images for validation.
    \item ImageNet-1K~\cite{deng2009imagenet} consists of 1,000 classes. The training and validation set contains 1,281,167 images and 50,000 images, respectively. We resize all images to the standard resolution 224$\times$224.
    \item ImageNet-21K-P~\cite{ridnik2021imagenet21k} removes infrequent classes from the original ImageNet-21K, resulting in 10,450 classes in total. There are 11,060,223 images for training and 522,500 images for validation. All images are resized to 224$\times$224 resolution.
    \item CIFAR-10~\cite{krizhevsky2009cifar} is a standard small-scale dataset consists of 60,000 32$\times$32 resolution images in 10 different classes. For each class, 5,000 images are used for training and 1,000 images are used for validation. 
    \item CIFAR-100~\cite{krizhevsky2009cifar} contains 100 classes. It has a training set with 50,000 images and a validation set with 10,000 images. 
\end{itemize}

\input{tables/tab-main}
\input{figures/cross-in1k}

In line with previous research practices~\cite{zhao2021dc, cazenavette2022mtt}, we employ the ``Images Per Class" (IPC) metric to indicate the capacity of the synthetic dataset. To uphold the diversity of the synthetic data, the number of curriculum stages, denoted as $J$, should grow in tandem with the increase in IPC. To strike a balance between scalability and effectiveness, we configure the growth of $J$ to follow a logarithmic increase with respect to IPC. More precisely, we define $J$ as follows: $J = \max(0, \lfloor log_{2}(\frac{\text{IPC}}{5}) \rfloor)+1$. For the hyperparameters specified in Equation~\eqref{eq:main}, we set $\alpha_{\text{reg}} = \alpha_{\text{adv}} = 1$. Regarding the proxy neural network architecture for dataset distillation, we generally employ ResNet-18 unless otherwise specified, consistent with the architecture used in SRe$^2$L. For Tiny-ImageNet, the first 7$\times$7 Conv layer of ResNet-18 is replaced by a 3$\times$3 Conv layer and the maxpool layer is discarded, following~\cite{yin2023sre2l, he2020moco}. We train 3 randomly initialized evaluation networks on the synthetic dataset to obtain the average accuracy and the error bar. We follow the same teacher model training protocol provided by~\cite{yin2023sre2l, yin2023cda}.

\subsection{Quantitative Comparisons}
\label{subsec:comparisons}
\textbf{Results on Large-Scale Datasets.} 
As SRe$^{2}$L~\cite{yin2023sre2l} and CDA~\cite{yin2023cda} are the first methods that achieve satisfactory performance on ImageNet-1K and ImageNet-21K respectively, we strictly follow its experimental setup and conduct comprehensive comparisons with it. The experimental results are detailed in Table~\ref{tab:main} and can be summarized as follows: 1) Our method \textbf{consistently} outperforms SRe$^{2}$L across all evaluated models, IPC settings, and evaluated datasets, achieving average improvements of 11.1\% on Tiny-ImageNet, 9.0\% on ImageNet-1K, and 7.3\% on ImageNet-21K. 2) SRe$^{2}$L, CDA, and our approach all demonstrate robust generalization ability beyond the specific network architecture used in the distillation process.

\input{tables/tab-cross-tiny}
Our method demonstrates more significant generalization abilities compared to SRe$^{2}$L on heterogeneous architectures. We conduct experiments across 10 heterogeneous architectures on ImageNet-1K, including DeiT-Tiny~\cite{touvron2021deit}, Swin-Tiny~\cite{liu2021swin}, ConvNeXt-Tiny~\cite{liu2022convnext}, DenseNet-121~\cite{huang2017densenet}, EfficientNet-V2~\cite{tan2021efficientnetv2}, MobileNet-V2~\cite{sandler2018mobilenetv2}, RegNet~\cite{radosavovic2020regnet}, ShuffleNet-V2~\cite{ma2018shufflenetv2}, ResMLP~\cite{touvron2022resmlp}, and MLP-Mixer~\cite{tolstikhin2021mlpmixer}.
The experimental outcomes are illustrated in Figure~\ref{fig:cross-in1k} and can be encapsulated as follows: 1) Our method realizes a \textbf{doubling} of performance on DeiT-Tiny and MLP-Mixer, marking an average improvement of 15.1\% over SRe$^{2}$L. 2) In the majority of network architectures, our performance remains competitive with that of ResNet-18, which has an accuracy of 57.4\%. 
Since previous mainstream dataset distillation methods~\cite{zhao2023dm, cazenavette2022mtt, kim2022idc, liu2023dream, guo2023datm} primarily focus on small datasets, and can hardly handle large-scale datasets, we compare our method with them on Tiny-ImageNet~\cite{le2015tiny}. For fairness, we adopt the 4-depth ConvNet architecture during the distillation stage for all the compared methods and set the multi-formation factors~\cite{kim2022idc} to 1 for both IDC~\cite{kim2022idc} and DREAM ~\cite{liu2023dream}.
As shown in Table~\ref{tab:cross-tiny}, our method achieves state-of-the-art performance across a variety of architectures~\cite{he2016resnet, huang2017densenet, tan2019efficientnet, sandler2018mobilenetv2, radosavovic2020regnet} among these approaches.

\textbf{Robustness to Corruptions.} 
\input{figures/robustness}
Our adversarial training not only enhances the primary objectives but also offers supplementary advantages. We evaluate the effectiveness of models trained on synthetic datasets in \textbf{out-of-domain} scenarios~\cite{kim2022bpc, liu2022haba, wei2023speed}. To assess the robustness of our models, we specifically employ ImageNet-C~\cite{hendrycks2018imagenet-c}, which encompasses a variety of datasets each corrupted by 19 distinct types of perturbations. Figure~\ref{fig:robustness} presents the mean accuracy across all 19 corruption types at two distinct corruption levels. In comparison to SRe$^{2}$L, our method demonstrates superior performance under \textbf{all kinds of corruptions}.

\textbf{Application in Continual Learning.} 
\input{figures/cl}
As a widespread application scenario for dataset distillation~\cite{zhao2023dm, liu2022haba, deng2022rtp, wei2023speed}, class-continual learning can reflect the average information content and diversity of each class in the synthetic dataset. We conduct continual learning experiments on ImageNet-1K with IPC-10 and use ResNet-18 and DenseNet-121 for evaluation. We randomly divide the 1000 classes into 5 learning steps, i.e., 200 classes per step. As illustrated in Figure~\ref{fig:cl}, CUDD consistently achieves the highest test accuracy at every stage for both evaluation models, highlighting its advantage over SRe$^2$L.

\input{tables/tab-small}
\input{figures/real-syn}
\input{tables/tab-mtt-ours}
\input{tables/tab-ablation}

\textbf{Results on Small-Scale Datasets.} 
Table~\ref{tab:small-supp} lists the evaluation results of different methods on networks involved in their distillation. The significance of these results is constrained, because dataset distillation methods tend to overfit on networks they used in distillation, leading to inflated performance metrics. It's crucial to acknowledge that the results of different network architectures can not be directly compared.

\textbf{Adaptation to Other Distillation Objectives.} 
Our method primarily provides a framework to generate synthetic images in separate sequential curricula to improve diversity without introducing significant computational overhead. Theoretically, any dataset distillation objective can be applied within each curriculum. However, a major objective of this paper is to scale up to large-scale datasets. Therefore, we adopt SRe$^2$L as our primary baseline, as it is the first to achieve satisfactory performance on large-scale datasets. To support this claim, we conduct an experiment where we apply our framework to MTT~\cite{cazenavette2022mtt} on Tiny-ImageNet with IPC 50. The results of these experiments are presented in Table.~\ref{tab:mtt-ours}. In this particular setup, synthetic images in each curriculum are optimized by MTT within our proposed framework, and the images are generated sequentially through multiple curricula. As evidence, this combination results in improved cross-architecture performance compared to applying MTT alone.

\subsection{Ablation Studies}
\label{subsec:ablation}
\textbf{Ablations on Objective Function.} 
Our ablation study, focusing on the objective function detailed in Equation~\eqref{eq:main}, is presented in Table~\ref{tab:reg-adv}. The results clearly demonstrate that incorporating both regularization loss and adversarial loss individually enhances performance. Moreover, the synergistic integration of these two losses yields an even more significant performance improvement. 

We also undertake a detailed comparative study on the application of regularization loss, examining its effectiveness in the pixel space as opposed to the feature space. The experimental results shown in Table~\ref{tab:pixel-feature} suggest minimal disparities between these two methodologies. 
% Further ablation studies are provided in the supplementary material for comprehensive insight.

\textbf{Ablations on Real-Image Initialization.} 
\input{tables/tab-init}
Furthermore, we perform ablation of real-image initialization on ImageNet-1K, including three scenarios: 1) SRe$^2$L with real-image initialization, 2) our method without any involvement of real images, and 3) our method with the regularization term $\mathcal{L}_\text{reg}$ disabled. As shown in Table~\ref{tab:init}, initialization from real images can bring improvements to SRe$^2$L, but there is still a gap between it and our method. In addition, as shown in Figure~\ref{fig:real-syn}, despite utilizing real images as a form of regularization in Equation~\eqref{eq:main}, the synthetic images exhibit marked differences from their real-image counterparts, effectively preserving the privacy of the original images, similar to previous dataset distillation methods~\cite{zhao2023dm, kim2022idc, cui2022dcbench, liu2023dream, guo2023datm}, which also adopt real image initialization. Moreover, our findings indicate that CUDD tends to produce objects of diverse scales within a single image, thereby enhancing the information density.

\textbf{Hyper-parameter Sensitivity Analysis.} 
\input{tables/tab-sensitivity}
We conduct experiments on the ImageNet-1K dataset using the IPC-10 configuration to assess the sensitivity of the hyper-parameters $\alpha_{\text{adv}}$ and $\alpha_{\text{reg}}$, as indicated in Table~\ref{tab:adv} and Table~\ref{tab:reg}, respectively. It is evident that these two hyper-parameters exhibit robust performance across a wide range and are not overly sensitive. Consequently, we opt for moderate values in our subsequent experiments.

\textbf{Effect of Adversarial Loss.} 
\input{tables/tab-constraint}
\input{figures/efficiency}
\input{figures/tsne}
To validate the effect of our adversarial loss, we employ t-SNE to visualize the distributions of synthetic data learned with and without this loss, alongside the original dataset, as shown in Figure~\ref{fig:tsne}. We use ResNet-18 to extract features for the deer class from CIFAR-10. The synthetic data contains 50 images, and the original data contains 5,000 images. As illustrated, synthetic data generated with the adversarial loss better covers the distribution of the original dataset, indicating that it retains richer information. We also try to allow the existence of the adversarial term regardless of whether the teacher can correctly recognize the current data, as shown in Table~\ref{tab:constraint}. We find that applying adversarial constraints can demonstrate greater advantages under small storage budgets, as synthetic data patterns are relatively easy to classify.

\subsection{Effectiveness and Efficiency of Curriculum Learning}
In comparison to SRe\(^{2}\)L, the training time overhead in our approach arises from the requirement to train a student network at the end of each curriculum, utilizing all synthetic images that have been generated up to that juncture. We introduce two strategies aimed at reducing the training cost, thereby enhancing the scalability of our method. 

\textbf{Logarithmic Curriculum Scheduling.} 
The first strategy, termed logarithmic curriculum scheduling, aims to decrease the overall number of curricula. As illustrated in Figure~\ref{fig:curricula-division}~(a), uniform curriculum scheduling, characterized by a dense series of curricula (IPC-\{5, 10, 15, 20, 25, 30, 35, 40\}), yields only a marginal improvement (1.6\% at IPC-40) compared to the logarithmic curriculum scheduling, which employs a sparser sequence of curricula (IPC-\{5, 10, 20, 40\}). However, this slight enhancement comes at the cost of doubling the training time (Figure~\ref{fig:curricula-division}~(b)). Therefore, we adopt the more cost-effective logarithmic curriculum scheduling.

\textbf{Curriculum Training.} 
The second strategy, termed curriculum training, is designed to reduce the number of training iterations required by the student network. Rather than initiating the student network's training from scratch for each curriculum, we employ the optimized parameters of the student network from the preceding curriculum as the initialization point, i.e., $\phi_j \leftarrow \phi_{j-1}^*$. Figure~\ref{fig:curriculum-student} depicts the results on CIFAR-10 under IPC-20 and IPC-50 configurations. Experimental results demonstrate that curriculum training is capable of converging to an accuracy comparable to that achieved by training from scratch, but requiring only \textbf{half} the iterations. Furthermore, as the curriculum progresses, the student network's capabilities steadily improve, enabling it to provide more valuable feedback during the subsequent stages of the curriculum.

\input{tables/tab-costs}
\textbf{Training Overheads.} We finally conduct experiments to compare the training costs between our method and SRe2L~\cite{yin2023sre2l}. We test the time consumed in the pre-classification stage (i.e. identifying samples misclassified by the previous student network and correctly classified by the teacher) before the data recovery stage and the total time for training the student network. These results are concluded in Table~\ref{tab:costs}. As shown, the student training accounts for a relatively large proportion of the overhead, and we have introduced logarithmic scheduling and curriculum training strategies to speed up this stage.

%% file: tables/tab-main.tex
\begin{table*}[!t]
\centering
    \caption{Comparisons on large-scale datasets. CUDD achieves state-of-the-art performance across all evaluated neural network architectures, under various images per class (IPC) configurations, and across all evaluated datasets. $^{\ast}$ denotes results obtained using the official code, while other results are directly taken from the original papers~\cite{yin2023sre2l, yin2023cda} We remove our original error bars since \cite{yin2023cda} does not provide all results with error bars}.
    \label{tab:main}
\resizebox{0.85\linewidth}{!}{
\begin{NiceTabular}{c|c|cc|cccc|cc}
    % \toprule
    & Dataset & \multicolumn{2}{c|}{Tiny-ImageNet} & \multicolumn{4}{c|}{ImageNet-1K} & \multicolumn{2}{c}{ImageNet-21K} \\
    \midrule[1pt]
    Architecture & IPC & 50 & 100 & 10 & 50 & 100 & 200 & 10 & 20 \\
    \midrule
    \multirow{3}{*}{ResNet-18} & SRe$^{2}$L~\cite{yin2023sre2l} & 41.1 & 49.7 & 21.3 & 46.8 & 52.8 & 57.0 & 19.3$^{*}$ & 21.6$^{*}$ \\
    & {CDA~\cite{yin2023cda}} & {48.7} & {53.2} & {33.5} & {53.5} & {58.0} & {63.3} & {22.6} & {26.4} \\
    & \textbf{CUDD} (Ours) & \cellcolor{gray!10}\textbf{55.6}  & \cellcolor{gray!10}\textbf{56.8} &
    \cellcolor{gray!10}\textbf{39.0} & \cellcolor{gray!10}\textbf{57.4} & \cellcolor{gray!10}\textbf{61.3} & \cellcolor{gray!10}\textbf{65.0} & \cellcolor{gray!10}\textbf{28.0} & \cellcolor{gray!10}\textbf{34.9} \\
    \midrule
    \multirow{3}{*}{ResNet-50} & SRe$^{2}$L~\cite{yin2023sre2l} & 42.2 & 51.2 & 28.4 & 55.6 & 61.0 & 64.6 & 28.6$^{*}$ & 30.5$^{*}$ \\
    & {CDA~\cite{yin2023cda}} & {49.7} & {54.4} & - & {61.3} & {65.1} & {67.6} & {32.4} & {35.3} \\
    & \textbf{CUDD} (Ours) & \cellcolor{gray!10}\textbf{57.0}  & \cellcolor{gray!10}\textbf{58.8} & 
    \cellcolor{gray!10}\textbf{46.2} & \cellcolor{gray!10}\textbf{63.6} & \cellcolor{gray!10}\textbf{66.7} & \cellcolor{gray!10}\textbf{68.6} & \cellcolor{gray!10}\textbf{34.1} & \cellcolor{gray!10}\textbf{36.1} \\
    \midrule
    \multirow{3}{*}{ResNet-101} & SRe$^{2}$L~\cite{yin2023sre2l} & 42.5 & 51.5 & 30.9 & 60.8 & 62.8 & 65.9 & 29.0$^{*}$ & 32.5$^{*}$ \\
    & {CDA~\cite{yin2023cda}} & {50.6} & {55.0} & - & {61.6} & {65.9} & {68.4} & {34.2} & {36.1} \\
    & \textbf{CUDD} (Ours) & \cellcolor{gray!10}\textbf{57.4}  & \cellcolor{gray!10}\textbf{59.4} & 
    \cellcolor{gray!10}\textbf{46.8} & \cellcolor{gray!10}\textbf{64.9} & \cellcolor{gray!10}\textbf{67.2} & \cellcolor{gray!10}\textbf{69.0} & \cellcolor{gray!10}\textbf{35.4} & \cellcolor{gray!10}\textbf{36.9} \\
    % \bottomrule
    \end{NiceTabular}
   }
    % \vspace{-1em}
\end{table*}

%% file: figures/cross-in1k.tex
\begin{figure*}[t]
    \centering
    \includegraphics[width=\linewidth]{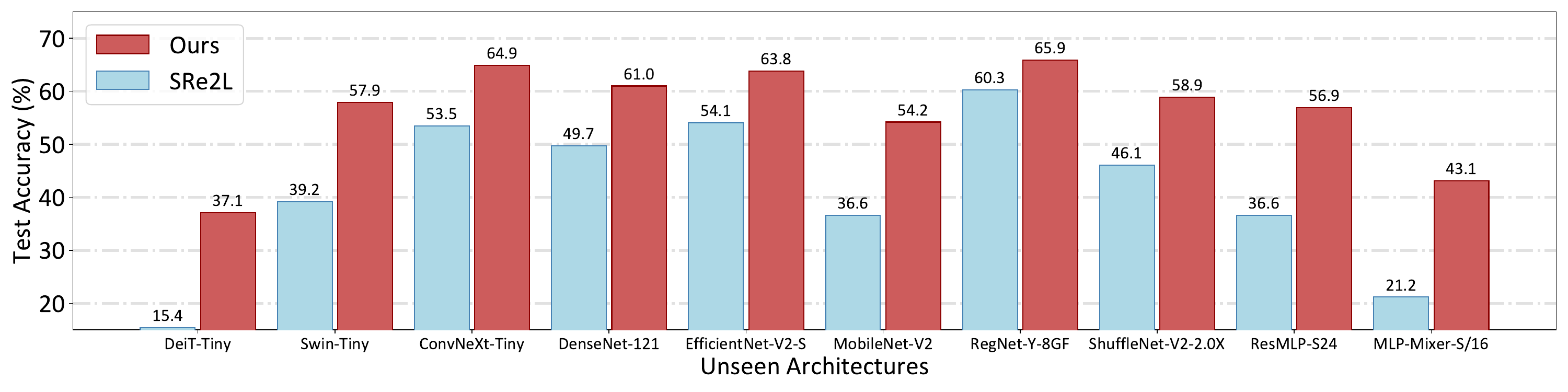}
    % \vspace{-20pt}
    \caption{Comparisons on heterogeneous architectures. CUDD achieves state-of-the-art performance on all 10 heterogeneous architectures with substantial leads. Experiments are conducted on ImageNet-1K using 50 images per class.}
    \label{fig:cross-in1k}
% \vspace{-.5em}
\end{figure*}

%% file: tables/tab-cross-tiny.tex
\begin{table*}[t]
\small
\centering
\caption{Cross-architecture performance on Tiny-ImageNet with 50 images per class. All the methods adopt 4-depth ConvNet for the distillation stage for fair comparisons. CUDD adheres to the protocol of SRe2L, which encompasses relabeling and knowledge distillation processes.}\
\label{tab:cross-tiny}
\resizebox{\linewidth}{!}{
\begin{NiceTabular}{c|cccccc|c}
% \toprule
Method & ConvNet-4 & ResNet-18 & DenseNet-121 &	RegNet-Y-800MF &	MobileNet-V2 & EfficientNet-B0 & Avg. \\
\midrule[1pt]
DM~\cite{zhao2023dm} & 24.1\tiny$\pm$0.3 & 29.9\tiny$\pm$0.4 & 24.9\tiny$\pm$0.3 & 16.5\tiny$\pm$0.2 & 18.4\tiny$\pm$0.4 &	21.8\tiny$\pm$0.3 & 22.6 \\
MTT~\cite{cazenavette2022mtt} & 28.0\tiny$\pm$0.3 & 30.9\tiny$\pm$0.2 & 29.0\tiny$\pm$0.4 & 18.3\tiny$\pm$0.5 & 19.8\tiny$\pm$0.1 & 26.9\tiny$\pm$0.2 & 25.5 \\
IDC~\cite{kim2022idc} & 25.2\tiny$\pm$0.2 & 32.4\tiny$\pm$0.7 & 29.1\tiny$\pm$0.2 & 24.3\tiny$\pm$0.3 & 25.6\tiny$\pm$0.5 & 27.0\tiny$\pm$0.4 & 27.3 \\
DREAM~\cite{liu2023dream} & 25.6\tiny$\pm$0.4 & 32.9\tiny$\pm$0.2 & 29.6\tiny$\pm$0.5 & 25.1\tiny$\pm$0.4 & 26.2\tiny$\pm$0.4 & 27.0\tiny$\pm$0.3 & 27.7 \\
DATM~\cite{guo2023datm} & 39.7\tiny$\pm$0.3 & 43.6\tiny$\pm$0.2 & 40.1\tiny$\pm$0.4 & 36.3\tiny$\pm$0.3 & \textbf{35.4\tiny$\pm$0.3} & 37.8\tiny$\pm$0.2 & 38.8 \\
\textbf{CUDD} (Ours) & \cellcolor{gray!10}\textbf{45.2\tiny$\pm$0.2} & \cellcolor{gray!10}\textbf{46.2\tiny$\pm$0.1} & \cellcolor{gray!10}\textbf{42.8\tiny$\pm$0.2} & \cellcolor{gray!10}\textbf{41.2\tiny$\pm$0.3} & \cellcolor{gray!10}\textbf{35.4\tiny$\pm$0.2} & \cellcolor{gray!10}\textbf{38.2\tiny$\pm$0.4} & \cellcolor{gray!10}\textbf{41.5}
 \\
\end{NiceTabular}
}
% \vspace{-5pt}
% \vspace{-1.5em}
\end{table*}

%% file: figures/robustness.tex
\begin{figure}[t]
\vspace{-1.5em}
\begin{minipage}[t]{0.5\linewidth}
    \centering
    \includegraphics[width=\linewidth]{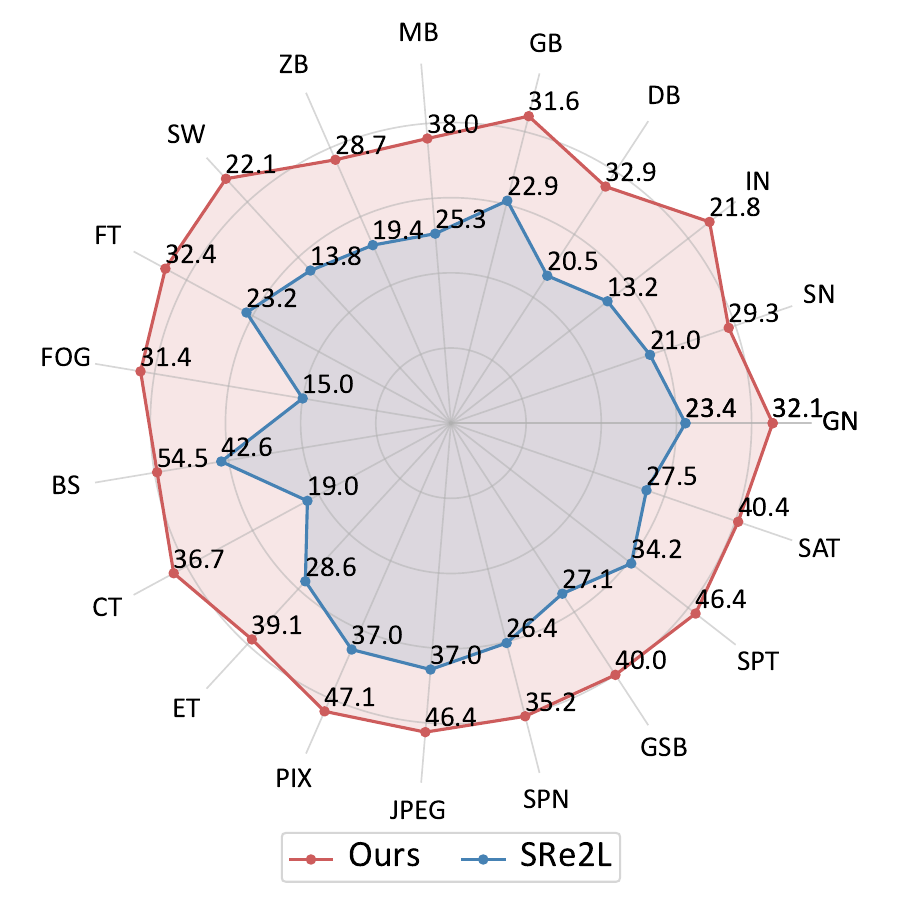}
    \centerline{(a) ResNet-18}
\end{minipage}%
\hfill
\begin{minipage}[t]{0.5\linewidth}
    \centering
    \includegraphics[width=\linewidth]{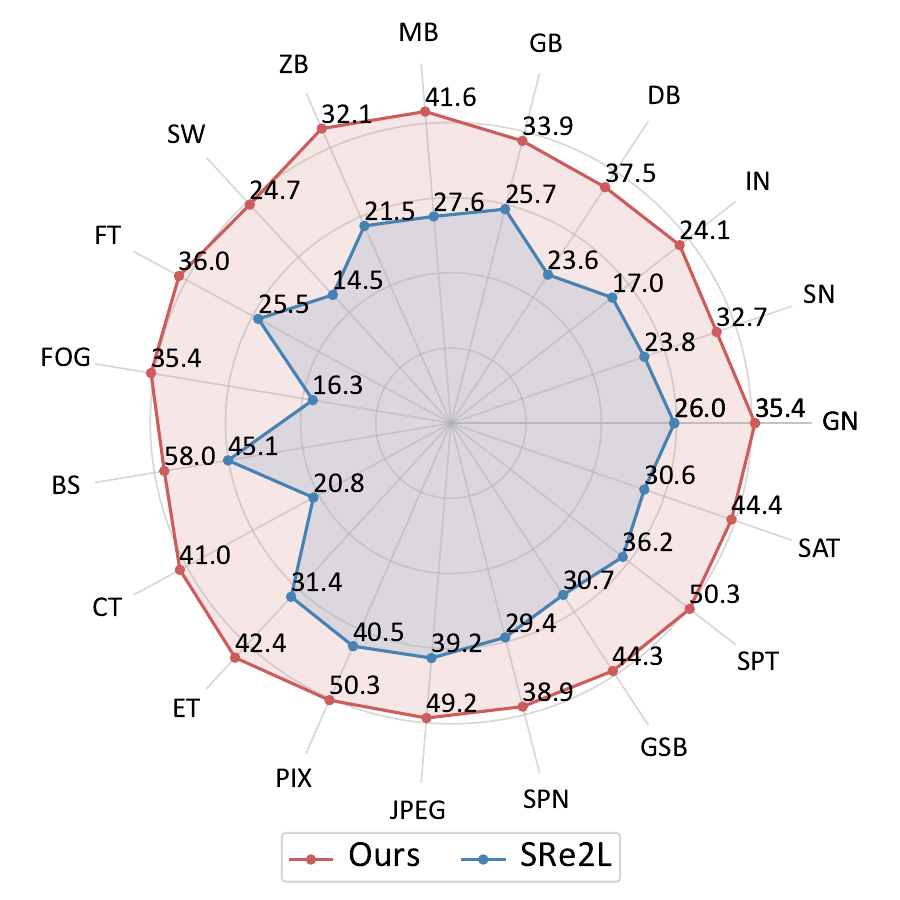}
    \centerline{(b) DenseNet-121}
\end{minipage}%
% \vspace{-5pt}
    \caption{Robustness against corruptions on ImageNet-C. Our method demonstrates enhanced robustness in two models trained on the synthetic dataset distilled from ImageNet-1K with IPC 50.}
    \label{fig:robustness}
% \vspace{-1.5em}
\end{figure}

%% file: figures/cl.tex
\begin{figure}[t]
\vspace{-1em}
\begin{minipage}[t]{0.48\linewidth}
    \centering
    \includegraphics[width=\linewidth]{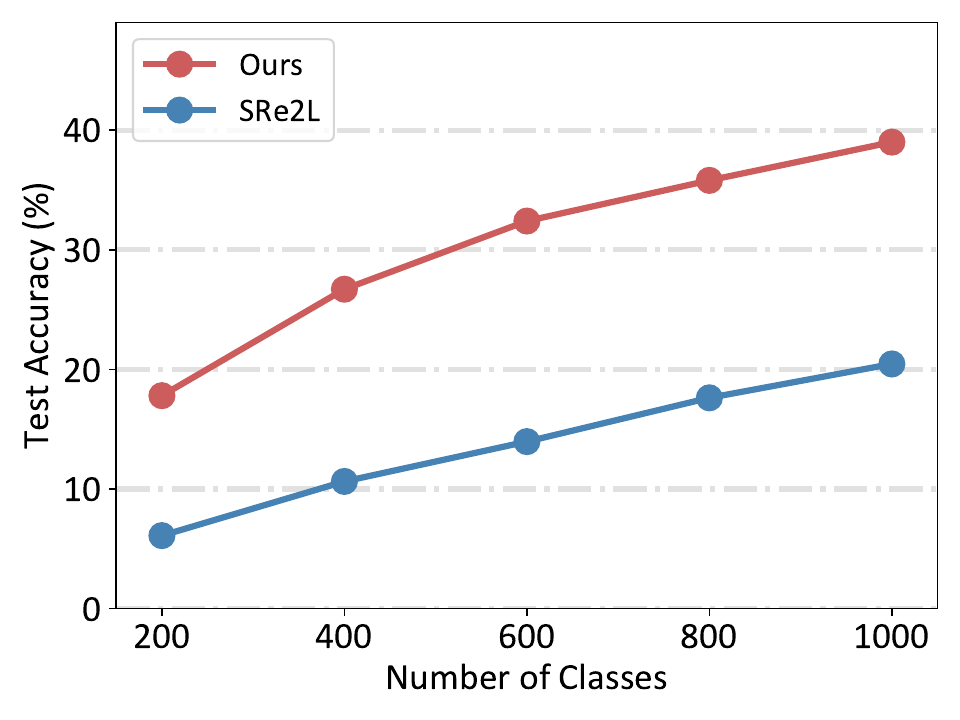}
    \centerline{(a) ResNet-18}
\end{minipage}%
\hfill
\begin{minipage}[t]{0.48\linewidth}
    \centering
    \includegraphics[width=\linewidth]{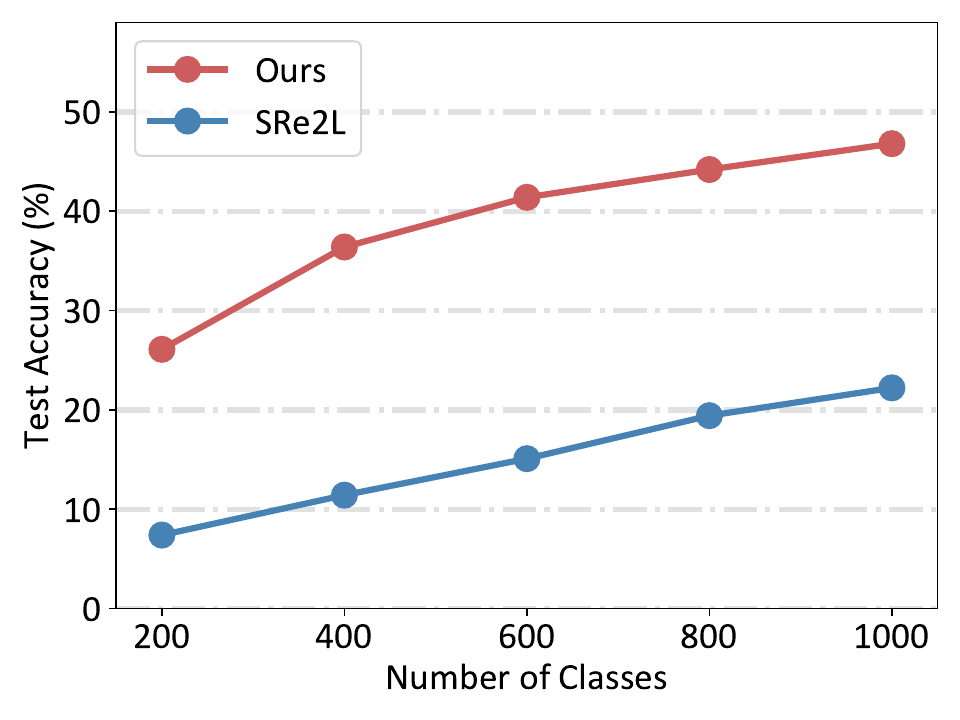}
    \centerline{(b) DenseNet-121}
\end{minipage}%
% \vspace{-5pt}
    \caption{Application in continual learning. On two different architectures, CUDD consistently maintains the highest test accuracy across all learning steps.}
    \label{fig:cl}
% \vspace{-.5em}
\end{figure}

%% file: tables/tab-small.tex
\begin{table}[t]
\centering
\caption{Comparisons with previous dataset distillation methods on small-scale datasets. C3-W128 denotes ConvNet-3-Width-128~\citep{zhao2021dc, cazenavette2022mtt}. R18 denotes ResNet-18. Evaluation is conducted on the same network structure that is involved in distilling. $^{\ast}$ denotes results obtained using the official code, and results of~\cite{zhao2021dc, zhao2021dsa, nguyen2021kip, zhao2023dm, cazenavette2022mtt} are taken from DC-BENCH~\cite{cui2022dcbench}, while other results are directly taken from their original papers. CUDD adheres to the protocol of SRe2L, which encompasses relabeling and knowledge distillation processes.}
\label{tab:small-supp}
\resizebox{\linewidth}{!}{
\begin{NiceTabular}{cl|cc|cc}
% \toprule
& & \multicolumn{2}{c}{CIFAR-10} & \multicolumn{2}{c}{CIFAR-100} \\
% \cmidrule{2-4}\cmidrule{5-7}\\
& Method $\setminus$ IPC & 10 & 50 & 10 & 50 \\
\midrule[1pt]
\multirow{12}{*}{\rotatebox[origin=c]{90}{C3-W128}} & DC~\cite{zhao2021dc} & 51.0\tiny$\pm$0.6 & 56.8\tiny$\pm$0.4 & 28.4\tiny$\pm$0.3 & 30.6\tiny$\pm$0.6 \\ 
& DSA~\cite{zhao2021dsa} & 53.0\tiny$\pm$0.4 & 60.3\tiny$\pm$0.4 & 32.2\tiny$\pm$0.4 & 43.1\tiny$\pm$0.3 \\ 
& KIP~\cite{nguyen2021kip} & 47.2\tiny$\pm$0.4 & 57.0\tiny$\pm$0.4 & 29.0\tiny$\pm$0.3 & - \\
& DM~\cite{zhao2023dm} & 47.6\tiny$\pm$0.6 & 62.0\tiny$\pm$0.3 & 29.2\tiny$\pm$0.3 & 42.3\tiny$\pm$0.4 \\
& MTT~\cite{cazenavette2022mtt} & 63.7\tiny$\pm$0.4 & 70.3\tiny$\pm$0.6 & 38.2\tiny$\pm$0.4 & 46.3\tiny$\pm$0.3 \\
& FRePo~\cite{zhou2022frepo} & 65.5\tiny$\pm$0.4 & 71.7\tiny$\pm$0.2 & 42.5\tiny$\pm$0.2 & 44.3\tiny$\pm$0.2 \\
& FTD~\cite{du2023ftd} & 66.6\tiny$\pm$0.3 & 73.8\tiny$\pm$0.2 & 43.4\tiny$\pm$0.3 & 50.7\tiny$\pm$0.3 \\
& TESLA~\cite{cui2023tesla} & 66.4\tiny$\pm$0.8 & 72.6\tiny$\pm$0.7 & 41.7\tiny$\pm$0.3 & 47.9\tiny$\pm$0.3 \\
& RCIG~\cite{loo2023rcig} & \textbf{69.1\tiny$\pm$0.4} & 73.5\tiny$\pm$0.3 & 44.1\tiny$\pm$0.4 & 46.7\tiny$\pm$0.3 \\
& DataDAM~\cite{sajedi2023datadam} & 54.2\tiny$\pm$0.8 & 67.0\tiny$\pm$0.4 & 34.8\tiny$\pm$0.5 & 49.4\tiny$\pm$0.3 \\
& SeqMatch~\cite{du2023seqmatch} & 66.2\tiny$\pm$0.6 & 74.4\tiny$\pm$0.5 & 41.9\tiny$\pm$0.5 & 51.2\tiny$\pm$0.3 \\
& DATM~\cite{guo2023datm} & 66.8\tiny$\pm$0.2 & \textbf{76.1\tiny$\pm$0.3} & 47.2\tiny$\pm$0.4 & 55.0\tiny$\pm$0.2 \\
& \textbf{CUDD} (Ours) & \cellcolor{gray!10}56.9\tiny$\pm$0.3 & \cellcolor{gray!10}72.7\tiny$\pm$0.2 & \cellcolor{gray!10}\textbf{49.0}\tiny$\pm$0.4 & \cellcolor{gray!10}\textbf{57.5}\tiny$\pm$0.2 \\
\midrule
\multirow{2}{*}{\rotatebox[origin=c]{90}{R18}} & SRe$^{2}$L~\cite{yin2023sre2l} & 27.8\tiny$\pm$0.5$^{*}$ & 48.9\tiny$\pm$0.6$^{*}$ & 23.5\tiny$\pm$0.8 & 51.4\tiny$\pm$0.8 \\
& \textbf{CUDD} (Ours) & \cellcolor{gray!10}\textbf{56.2\tiny$\pm$0.4} & \cellcolor{gray!10}\textbf{84.5\tiny$\pm$0.3} & \cellcolor{gray!10}\textbf{60.3\tiny$\pm$0.2} & \cellcolor{gray!10}\textbf{65.7\tiny$\pm$0.2} \\
% \bottomrule
\end{NiceTabular}
}
% \vspace{-2pt}
% \vspace{-.5em}
\end{table}

%% file: figures/real-syn.tex
\begin{figure*}[t]
    \centering
    \includegraphics[width=\linewidth]{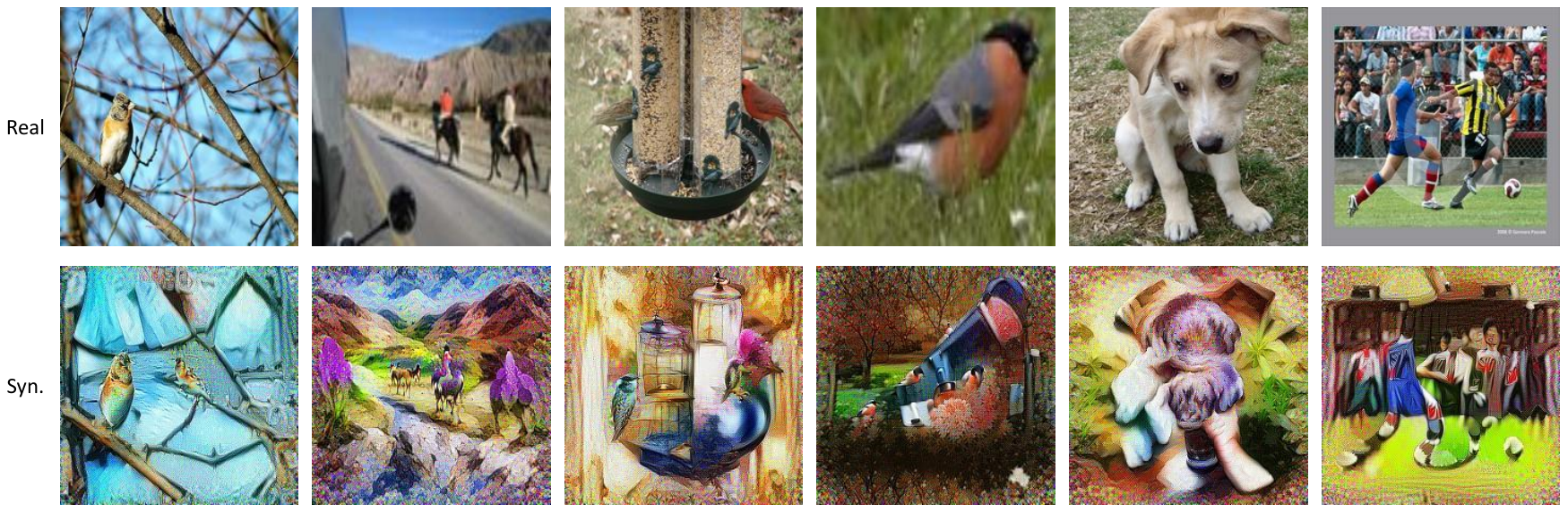}
\vspace{-15pt}
    \caption{Comparisons between real images employed as initial seeds and the corresponding optimized synthetic images.}
    \label{fig:real-syn}
\vspace{-.5em}
\end{figure*}

%% file: tables/tab-mtt-ours.tex
\begin{table*}[t]
\small
\centering
\caption{Adapting our curriculum framework to other distillation objectives.}
\label{tab:mtt-ours}
\resizebox{\linewidth}{!}{
\begin{NiceTabular}{c|cccccc|c}
% \toprule
Method & ConvNet-4 & ResNet-18 & DenseNet-121 & RegNet-Y-800MF &	MobileNet-V2 & EfficientNet-B0 & Avg. \\
\midrule[1pt]
MTT~\cite{cazenavette2022mtt} & 28.0\tiny$\pm$0.3 & 30.9\tiny$\pm$0.2 & 29.0\tiny$\pm$0.4 & 18.3\tiny$\pm$0.5 & 19.8\tiny$\pm$0.1 & 26.9\tiny$\pm$0.2 & 25.5 \\
CUDD (Ours) & \cellcolor{gray!10}45.2\tiny$\pm$0.2 & \cellcolor{gray!10}46.2\tiny$\pm$0.1 & \cellcolor{gray!10}42.8\tiny$\pm$0.2 & \cellcolor{gray!10}41.2\tiny$\pm$0.3 & \cellcolor{gray!10}35.4\tiny$\pm$0.2 & \cellcolor{gray!10}38.2\tiny$\pm$0.4 & \cellcolor{gray!10}41.5
 \\
MTT~\cite{cazenavette2022mtt} w/ our framework & \textbf{45.4\tiny$\pm$0.4} & \textbf{46.5\tiny$\pm$0.3} & \textbf{43.5\tiny$\pm$0.2} & \textbf{43.2\tiny$\pm$0.2} & \textbf{37.6\tiny$\pm$0.3} & \textbf{38.7\tiny$\pm$0.5} & \textbf{42.5}\\
\end{NiceTabular}
}
% \vspace{-5pt}
% \vspace{-.5em}
\end{table*}

%% file: tables/tab-ablation.tex
\begin{table*}[t]
\begin{minipage}{0.52\linewidth}
\small
\centering
\caption{Ablations on the objective function.}
\label{tab:reg-adv}
\resizebox{0.9\linewidth}{!}{
\begin{NiceTabular}{cc|ccc}
% \toprule
adv. & reg. & ResNet-18 & ResNet-50 & ResNet-101 \\
\midrule[1pt]
- & - & 34.4\tiny$\pm$0.2 & 39.1\tiny$\pm$0.3 & 42.9\tiny$\pm$0.3 \\
\checkmark & - & 36.0\tiny$\pm$0.3 & 41.6\tiny$\pm$0.5 & 45.3\tiny$\pm$0.4 \\
- & \checkmark & 38.3\tiny$\pm$0.2 & 44.7\tiny$\pm$0.3 & 44.9\tiny$\pm$0.2 \\
\textbf{\checkmark} & \textbf{\checkmark} & \cellcolor{gray!10}\textbf{39.0\tiny$\pm$0.4} & \cellcolor{gray!10}\textbf{46.2\tiny$\pm$0.6} & \cellcolor{gray!10}\textbf{46.8\tiny$\pm$0.3} \\
% \bottomrule
\end{NiceTabular}
}
% \vspace{-5pt}
% \vspace{-.5em}
\end{minipage}
\hfill
\begin{minipage}{0.47\linewidth}
\small
\centering
\caption{Comparative analysis of regularization loss applied in pixel space versus feature space.}
\label{tab:pixel-feature}
\resizebox{0.9\linewidth}{!}{
\begin{NiceTabular}{c|ccc}
% \toprule
space & ResNet-18 & ResNet-50 & ResNet-101 \\
\midrule[1pt]
\textbf{pixel} & \cellcolor{gray!10}\textbf{39.0\tiny$\pm$0.4} & \cellcolor{gray!10}\textbf{46.2\tiny$\pm$0.6} & \cellcolor{gray!10}46.8\tiny$\pm$0.3 \\
feature & 38.7\tiny$\pm$0.2 & 45.7\tiny$\pm$0.4 & \textbf{47.5}\tiny$\pm$0.4 \\
% \bottomrule
\end{NiceTabular}
}
% \vspace{-5pt}
% \vspace{-.5em}
\end{minipage}
% \vspace{-1em}
\end{table*}

%% file: tables/tab-init.tex
\begin{table}[t]
% \vspace{-1mm}
\small
\centering
\caption{Ablations on real-image initialization.}
\label{tab:init}
\resizebox{0.9\linewidth}{!}{
\begin{NiceTabular}{c|ccc|cc}
% \toprule
Method & {adv.} & init. & $\mathcal{L}_\text{reg}$ & IPC 10 & IPC 50 \\
\midrule[1pt]
\multirow{2}{*}{SRe$^2$L} & - & - & - & 21.3\tiny$\pm$0.6 & 46.8\tiny$\pm$0.2 \\
& - & \checkmark & - & 22.8\tiny$\pm$0.3 & 48.0\tiny$\pm$0.2 \\
\midrule
\multirow{3}{*}{\textbf{CUDD} (Ours)} & \checkmark & - & - & 33.5\tiny$\pm$0.3 & 54.8\tiny$\pm$0.1 \\
& \checkmark & \checkmark & - & 36.0\tiny$\pm$0.3 & 55.9\tiny$\pm$0.2 \\
& \checkmark & \checkmark & \checkmark & \cellcolor{gray!10}\textbf{39.0\tiny$\pm$0.4} & \cellcolor{gray!10}\textbf{57.4\tiny$\pm$0.2} \\
\end{NiceTabular}
}
% \vspace{-4pt}
% \vspace{-1em}
\end{table}

%% file: tables/tab-sensitivity.tex
\begin{table*}[t]
\begin{minipage}{0.49\linewidth}
\small
\centering
\caption{Sensitivity analysis of $\alpha_{\text{adv}}$.}
\label{tab:adv}
\resizebox{\linewidth}{!}{
\begin{NiceTabular}{c|ccc}
% \toprule
$\alpha_{\text{adv}}$ & ResNet-18 & ResNet-50 & ResNet-101 \\
\midrule[1pt]
0.3 & 38.9\tiny$\pm$0.2 & 45.0\tiny$\pm$0.4 & 46.7\tiny$\pm$0.3 \\
\textbf{1.0} & \cellcolor{gray!10}\textbf{39.0\tiny$\pm$0.4} & \cellcolor{gray!10}\textbf{46.2\tiny$\pm$0.6} & \cellcolor{gray!10}46.8\tiny$\pm$0.3 \\
3.0 & 38.8\tiny$\pm$0.4 & 44.0\tiny$\pm$0.5 & \textbf{47.3\tiny$\pm$0.4} \\
% \bottomrule
\end{NiceTabular}
}
% \vspace{-5pt}
% \vspace{-.5em}
\end{minipage}
\hfill
\begin{minipage}{0.49\linewidth}
\small
\centering
\caption{Sensitivity analysis of $\alpha_{\text{reg}}$.}
\label{tab:reg}
\resizebox{\linewidth}{!}{
\begin{NiceTabular}{c|ccc}
% \toprule
$\alpha_{\text{reg}}$ & ResNet-18 & ResNet-50 & ResNet-101 \\
\midrule[1pt]
0.3 & 38.3\tiny$\pm$0.2 & 45.3\tiny$\pm$0.5 & 46.1\tiny$\pm$0.4 \\
\textbf{1.0} & \cellcolor{gray!10}\textbf{39.0\tiny$\pm$0.4} & \cellcolor{gray!10}\textbf{46.2\tiny$\pm$0.6} & \cellcolor{gray!10}\textbf{46.8\tiny$\pm$0.3} \\
3.0 & 37.9\tiny$\pm$0.2 & 45.0\tiny$\pm$0.4 & 45.7\tiny$\pm$0.4 \\
% \bottomrule
\end{NiceTabular}
}
% \vspace{-5pt}
% \vspace{-.5em}
\end{minipage}
\end{table*}

%% file: tables/tab-constraint.tex
\begin{table*}[t]
% \vspace{-4mm}
% \begin{table*}
% \small
\centering
\caption{Effect of adversarial constraint.}
\label{tab:constraint}
\resizebox{0.75\linewidth}{!}{
\begin{NiceTabular}{c|ccc|ccc}
% \toprule
& \multicolumn{3}{c}{CIFAR-10} & \multicolumn{3}{c}{CIFAR-100} \\
IPC & 10 & 20 & 50 & 10 & 20 & 50 \\
\midrule[1pt]
none & 53.2\tiny$\pm$0.4 & 73.6\tiny$\pm$0.4 & 84.0\tiny$\pm$0.3 & 59.7\tiny$\pm$0.4 & 63.0\tiny$\pm$0.2 & 65.4\tiny$\pm$0.3 \\
\textbf{constraint} & \cellcolor{gray!10}\textbf{56.2\tiny$\pm$0.4} & \cellcolor{gray!10}\textbf{74.3\tiny$\pm$0.2} & \cellcolor{gray!10}\textbf{84.5\tiny$\pm$0.3} & \cellcolor{gray!10}\textbf{60.3\tiny$\pm$0.2} & \cellcolor{gray!10}\textbf{63.5\tiny$\pm$0.3} & \cellcolor{gray!10}\textbf{65.7\tiny$\pm$0.2}\\
% \bottomrule
\end{NiceTabular}
}
% \vspace{-5pt}
% \vspace{-.5em}
\end{table*}

%% file: figures/efficiency.tex
\begin{figure*}[t]
\begin{minipage}[t]{0.49\linewidth}
\begin{minipage}[t]{0.49\linewidth}
    \centering
    \includegraphics[width=\linewidth]{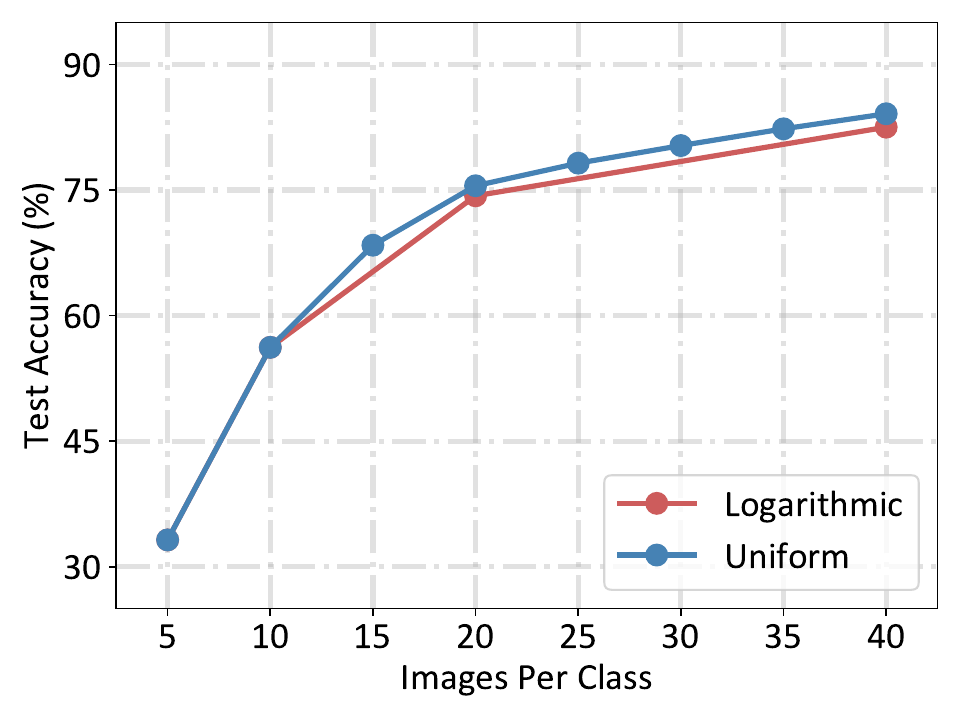}
    % \vspace{-12pt}
    \centerline{(a) Test Accuracy}
\end{minipage}
\hfill
\begin{minipage}[t]{0.49\linewidth}
    \centering
    \includegraphics[width=\linewidth]{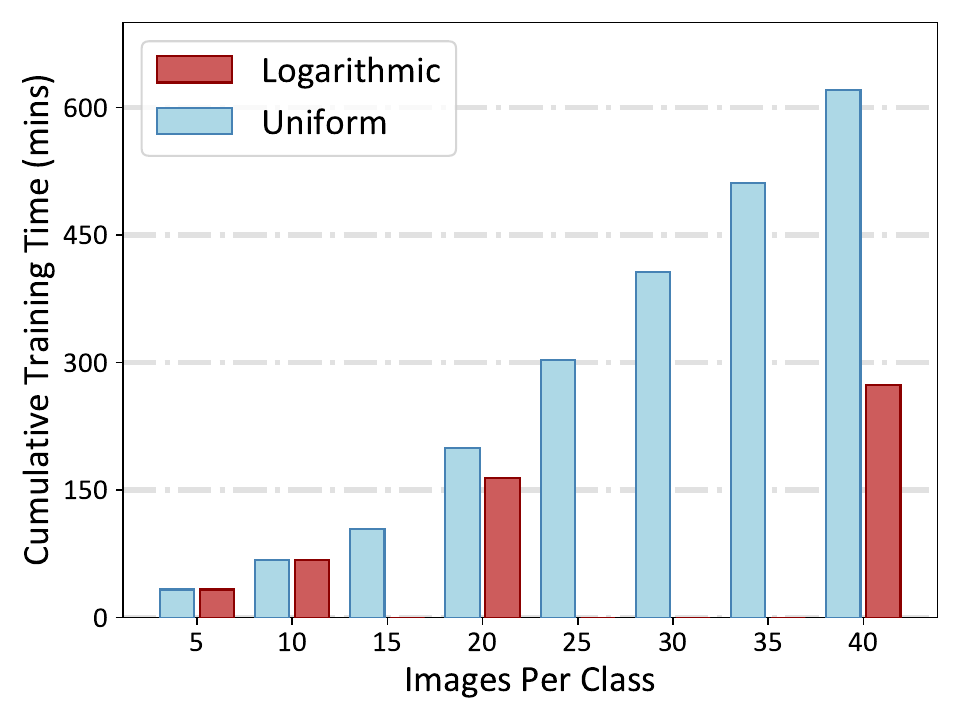}
    % \vspace{-12pt}
    \centerline{(b) Cumulative Training Time}
\end{minipage}\\
% \vspace{-5pt}
\caption{Comparative analysis of logarithmic curriculum scheduling versus uniform curriculum scheduling. Results for the logarithmic curriculum scheduling strategy are presented only for IPC values of \{5, 10, 20, 40\}. Consequently, despite the uniformly spaced axis ticks, no data is shown for this strategy at IPCs \{15, 25, 30, 35\}. Experiments are conducted on CIFAR-10 using a single RTX 3090 GPU.} 
\label{fig:curricula-division}
% \vspace{-.5em}
\end{minipage}
\hfill
\begin{minipage}[t]{0.49\linewidth}
\begin{minipage}[t]{0.49\linewidth}
    \centering
    \includegraphics[width=\linewidth]{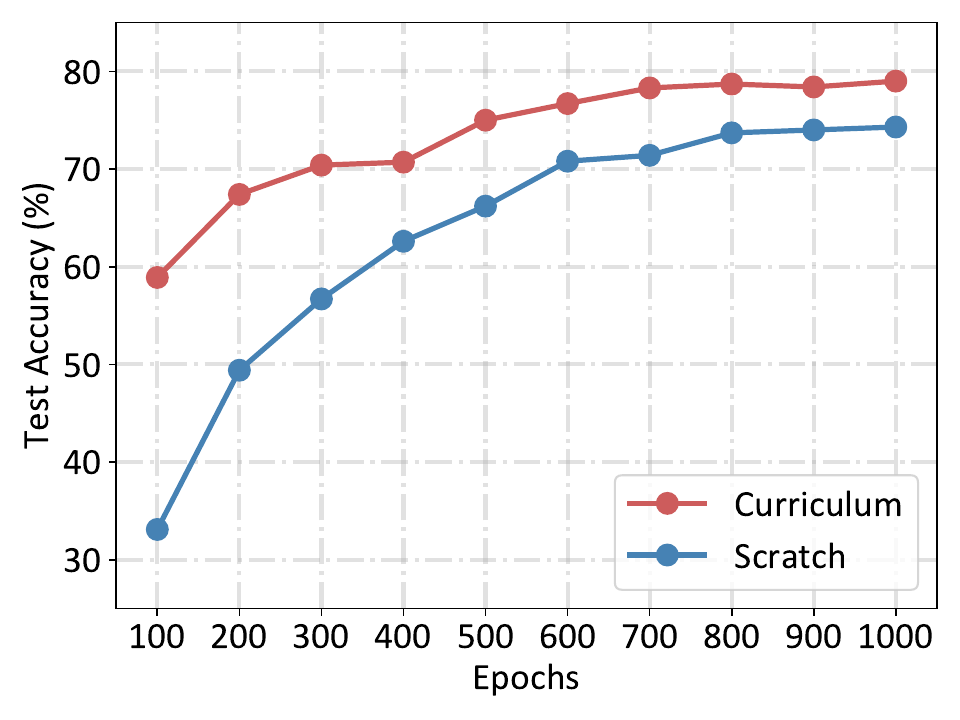}
    \centerline{(a) IPC-20}
\end{minipage}%
\hfill
\begin{minipage}[t]{0.49\linewidth}
    \centering
    \includegraphics[width=\linewidth]{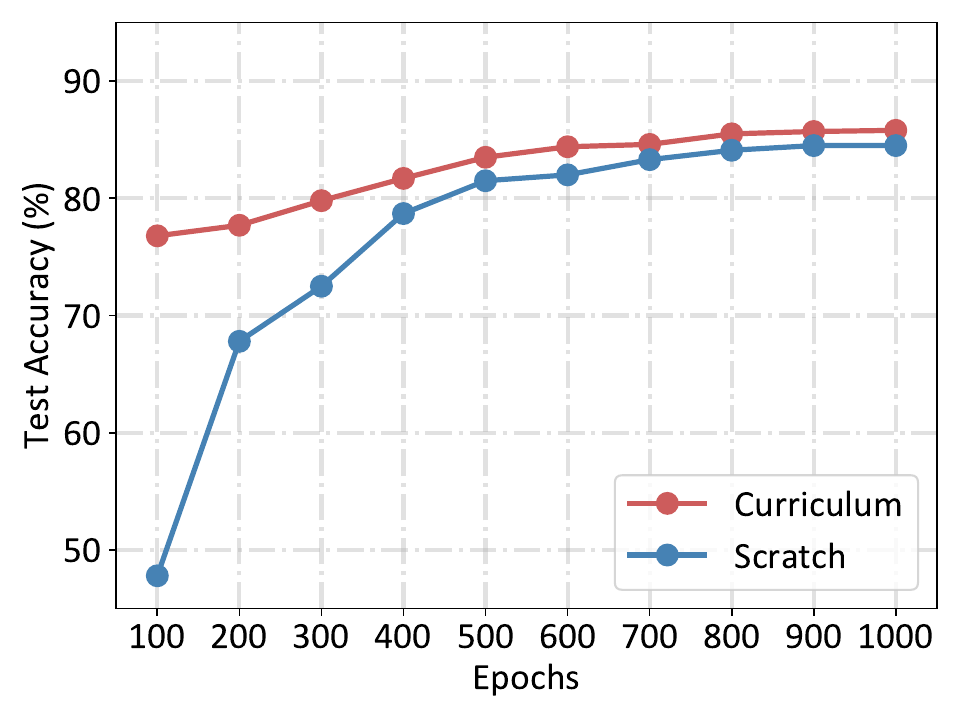}
    \centerline{(b) IPC-50}
\end{minipage}%
% \vspace{-5pt}
\caption{Comparison of student network training iterations: curriculum training (initializing from the previous curriculum) versus training from scratch. The student network continuously improves through the curricula.}
\label{fig:curriculum-student}
% \vspace{-.5em}
\end{minipage}
% \vspace{-1em}
\end{figure*}

%% file: figures/tsne.tex
\begin{figure}[t]
\vspace{-1em}
\begin{minipage}[t]{0.48\linewidth}
    \centering
    \includegraphics[width=\linewidth]{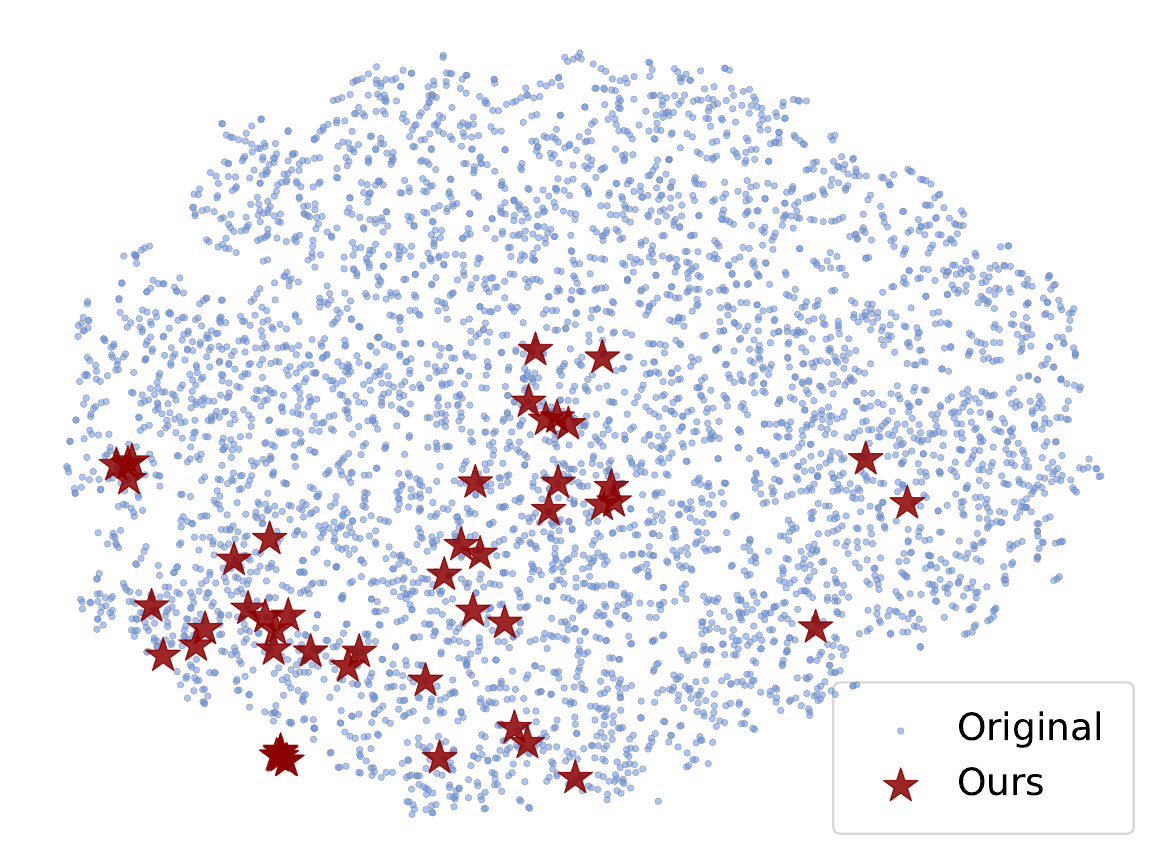}
    \centerline{(a) w/o adv.}
\end{minipage}%
\hfill
\begin{minipage}[t]{0.48\linewidth}
    \centering
    \includegraphics[width=\linewidth]{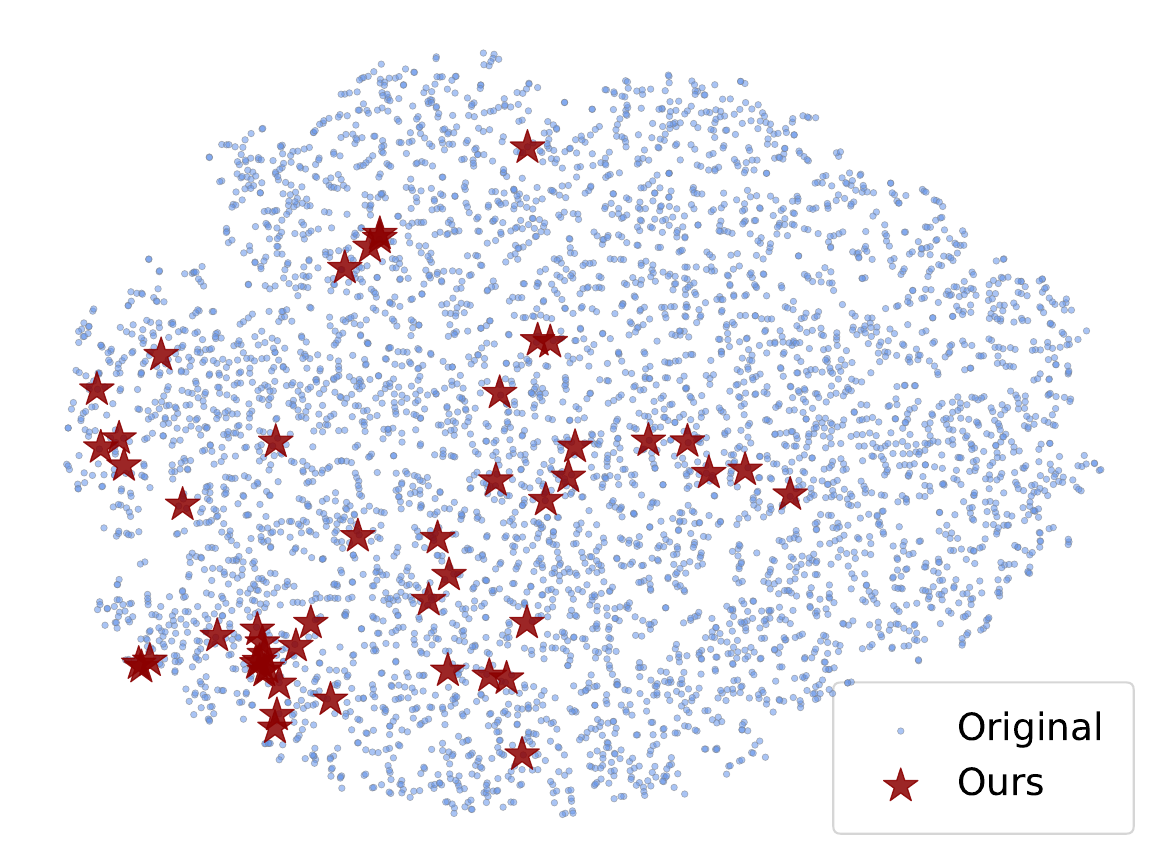}
    \centerline{(b) w/ adv.}
\end{minipage}%

\caption{Distributions of the deer class of CIFAR-10. We compare synthetic data (50 images) learned with and without adversarial loss (adv.) to the original dataset (5,000 images). ResNet-18 is used for feature extraction. The adversarial loss promotes synthetic data to achieve better coverage of the original dataset's distribution and retain higher informativeness.}
\label{fig:tsne}
\end{figure}

%% file: tables/tab-costs.tex
\begin{table}[t]
% \vspace{-1mm}
\small
\centering
\caption{Training cost comparisons on CIFAR-10 with IPC 10 (mins). Cls. denotes the pre-classification stage. Student denotes the student network training.}
\label{tab:costs}
\resizebox{\linewidth}{!}{
\begin{NiceTabular}{c|ccc|cc}
% \toprule
Method & Recover & Relabel & Training & Cls. & Student \\
\midrule[1pt]
SRe$^2$L & 6.0m & 3.7m & 78.8m & - & - \\
% \midrule
\textbf{CUDD} (Ours) & \cellcolor{gray!10}6.1m & \cellcolor{gray!10}3.7m & \cellcolor{gray!10}78.8m & \cellcolor{gray!10}1.5m & \cellcolor{gray!10}40.2m \\
\end{NiceTabular}
}
% \vspace{-4pt}
% \vspace{-1em}
\end{table}

%% file: sections/related.tex
\section{Related Works}
\label{sec:related}
\paragraph{Dataset Distillation} 
Dataset distillation~\cite{wang2018dd, cui2022dcbench} aims to learn a compact synthetic dataset that captures crucial information from the original dataset. \cite{wang2018dd} first proposed a bi-level optimization framework to handle this task, further extended by \cite{deng2022rtp}. To alleviate the computationally intensive optimization procedure of this formulation, various surrogate objectives have been proposed, including KRR-based approaches~\cite{nguyen2021kip, zhou2022frepo, loo2022rfad, loo2023rcig}, parameter-based methods~\cite{zhao2021dc, zhao2021dsa, cazenavette2022mtt, lee2022dcc, kim2022idc, liu2022haba, du2023ftd, cui2023tesla, wei2023speed, du2023seqmatch, guo2023datm, lee2024selmatch, liu2024att}, and distribution-based techniques~\cite{zhao2023dm, wang2022cafe, zhao2022itgan, lee2022kfs, zhao2023idm, sajedi2023datadam}. 

Recent studies have begun to explore fully disentangled methods for large-scale dataset distillation~\cite{yin2023sre2l, yin2023cda, shao2023gvbsm, sun2023rded}, leveraging a trained teacher network to produce synthetic datasets. For instance, \cite{yin2023cda} proposes a method for more effective optimization of individual images. \cite{shao2023gvbsm} utilizes multiple teacher architectures for better generalization ability. \cite{sun2023rded} composes each synthetic image by cropping multiple real image patches without further optimization. Compared to these concurrent works, CUDD: 1) performs feedback evaluation on the original dataset to ensure better coverage of its distribution, and 2) considers inter-image diversity, utilizing a novel optimization method to further enhance it.

\paragraph{Coreset Selection and Data Augmentation} 
Coreset selection~\cite{welling2009herding, chen2010super, feldman2011scalable, rebuffi2017icarl} identifies a subset of the most representative samples from the original dataset. Methods like \cite{welling2009herding, sener2018kcenter} eliminate redundant samples based on their similarity to the remaining ones. Other approaches \cite{margatina2021cal, ducoffe2018deepfool} select samples based on their learning difficulty. Data augmentation~\cite{krizhevsky2012alexnet, zhang2018mixup, yun2019cutmix, zhao2020diffaug, karras2020ada} applies deterministic or random transformations to increase the diversity of the original data while preserving most of the original information. In contrast, CUDD directly optimizes the synthetic data, resulting in a high distortion rate of the original data and the ability to prevent the leakage of private information. Moreover, data augmentation and our method are complementary to each other.

\paragraph{Curriculum Learning} 
Curriculum learning~\cite{bengio2009curriculumlearning} is originally defined as a way to train networks by organizing the order in which data is fed to the network. Data sorting can be based on priori rules~\cite{bengio2009curriculumlearning}, model performance~\cite{kumar2010self, lee2011learning}, and data diversity~\cite{zhang2015self, soviany2020curriculum}. Leveraging the curriculum concept, \cite{morerio2017curriculumdropout} decreases the probability of dropout during training. \cite{sinha2020curriculumbysmoothing} proposes to gradually deblur convolutional activation maps. CUDD orchestrates curricula for data synthesis as well as student network training, thereby enriching data diversity while also enhancing the distillation efficiency.

\section{Concurrent Works}
Several concurrent works have explored various strategies to enhance dataset distillation. For instance, \cite{li2024pad} implicitly increases the difficulty of the curriculum by leveraging the complexity of the network, wherein subsets of the synthetic dataset are selected based on layers extracted from varying depths of the proxy models during the distillation process. \cite{wang2024edf} applies Grad-CAM activation maps to emphasize key discriminative regions while minimizing low-loss supervision signals, thereby mitigating the presence of common patterns in synthetic images. Real-image initialization is explored in \cite{shao2024edc} and \cite{du2024dwa}, demonstrating the advantages of using real image initialization in enhancing generalization performance and accelerating convergence. Feature integration across classes is addressed in \cite{zhang2025infer}, which generates multiple additional synthetic instances from a single universal feature compensator input. Additionally, \cite{yu2024teddy} introduces a memory-efficient approximation derived from Taylor expansion, transforming the original form dependent on multi-step gradients into a first-order one. In comparison, CUDD explicitly defines the difficulty of the curriculum through adversarial loss and selective initialization.

%% file: sections/conclusion.tex
\section{Conclusion, Broader Impacts, and Limitations}
\label{sec:conclusion}
\paragraph{Conclusion} In this paper, we propose CUDD, a simple and neat framework for curriculum dataset distillation. We decompose the synthesis of data into multiple curricula and utilize student networks for knowledge transmission between these curricula. First, we perform feedback evaluation on the previously trained student network, sampling an original subset to serve as initialization points of the synthetic subset. This ensures better coverage of the original data distribution. Second, we utilize both the teacher and student networks for adversarial optimization of the current synthetic subset. This approach enhances data representativeness and fosters better generalization across various network architectures. Finally, we train the student network to gain knowledge of existing synthetic data. We further introduce logarithmic curriculum scheduling and curriculum student training to accelerate the distilling process. Extensive experiments prove that our framework achieves state-of-the-art on various benchmarks.

\paragraph{Broader Impacts} CUDD markedly advances the feasibility of dataset distilling across extensive data scales, yielding notable academic and societal implications: 1) It advocates for eco-friendly AI advancements by minimizing both training expenses and data storage requirements. 2) It bolsters the safeguarding and dissemination of datasets within domains sensitive to privacy concerns. 3) It enriches the comprehension of the efficacy inherent in datasets.

\paragraph{Limitations} We acknowledge certain limitations in CUDD: 1) At present, our method does not achieve lossless dataset distillation, indicating a potential loss of information during the distillation process on large-scale datasets. 2) There is a possibility that biases present in the original dataset may be preserved or even amplified in the synthetic dataset, potentially leading to biased outcomes.

%% file: sections/acknowledgements.tex
\section*{Acknowledgements}
This work was supported in part by the National Natural Science Foundation of China (62206271), and the Fundamental Research Funds for the Central Universities (No. xxj032023020), and the Shenzhen Key Technical Projects under Grant (JSGG20220831105801004, CJGJZD20220517141605013, JCYJ20220818101406014), and the Characteristic Innovation Project of Ordinary Universities in Guangdong Province (2024KTSCX026), and the Guangdong Provincial Key Laboratory of Computility Microelectronics (2024B1212010007).

%% file: sections/appendix.tex
% \newpage
% \appendix

% \twocolumn
\section*{Supplemental Materials}
\label{sec:appendix}
% This appendix is structured as follows: 
% \begin{itemize}[topsep=1pt, partopsep=9pt, itemsep=-1pt, parsep=0.5ex]
%     % \item In Sec.~\ref{suppsec:datasets}, we provide details of the dataset.
%     % \item In Sec.~\ref{suppsec:results}, we present additional qualitative results accompanied by in-depth analyses.
%     \item In Sec.~\ref{suppsec:hyper-costs}, we conclude our detailed hyper-parameters and discuss the computational costs.
%     \item In Sec.~\ref{suppsec:visualizations}, we provide more visualizations.
% \end{itemize}

% \subsection{Dataset Details} 
% \label{suppsec:datasets}

% \subsection{Additional Results and Analyses} 
% \label{suppsec:results}

\subsection{Hyper-Parameters and Computational Costs} 
\label{suppsec:hyper-costs}
We detail the hyper-parameters for data synthesis and their subsequent evaluation in downstream model training within Table~\ref{tab:recover} and Table~\ref{tab:validation}, respectively. All of our experiments can be conducted on a single 24GB RTX 3090 GPU. The wall clock distilling time of IPC-1 and peak GPU memory costs are concluded as follows: \{35 seconds, 0.8 GB\} for CIFAR-10, and \{40 seconds, 2.1 GB\} for CIFAR-100, \{17 minutes, 6.3 GB\} for Tiny-ImageNet, \{86 minutes, 8.3 GB\} for ImageNet-1K, \{7.5 hours, 8.3 GB\} for ImageNet-21K. It is important to note that the distillation time is primarily influenced by factors such as the total number of classes, image resolution, and the number of iterations.
\input{tables/tab-recover}
\input{tables/tab-validation}
\input{figures/supp-compare}

\subsection{Additional Visualizations} 
\label{suppsec:visualizations}
Additional visualizations are provided in the subsequent section. Figure~\ref{fig:supp-compare}~(a)-(e) depict comparisons between our synthetic data and that generated by SRe$^2$L. Additionally, extensive collections of our synthetic data on ImageNet-1K and ImageNet-21K are displayed in Figure~\ref{fig:supp-vis-in1k} and Figure~\ref{fig:supp-vis-in21k}, respectively. 
\input{figures/supp-vis-in1k}
\input{figures/supp-vis-in21k}

%% file: tables/tab-recover.tex
\begin{table}[h]
\centering
\resizebox{\linewidth}{!}{
\begin{NiceTabular}{c|ccccc}
% \toprule
config & CIFAR-10 & CIFAR-100 & Tiny-ImageNet & ImageNet-1K & ImageNet-21K \\
\midrule[1pt]
optimizer & \multicolumn{5}{c}{Adam} \\
momentum & \multicolumn{5}{c}{$\beta_1$, $\beta_2$ = 0.5, 0.9}  \\
weight decay & \multicolumn{5}{c}{1e-4} \\
lr schedule & \multicolumn{5}{c}{cosine} \\
augmentation & \multicolumn{5}{c}{Random Resized Crop}  \\
$\alpha_{\text{adv}}$ & \multicolumn{5}{c}{1.0}  \\
$\alpha_{\text{reg}}$ & \multicolumn{5}{c}{1.0}  \\
learning rate & 0.25 & 0.25 & 0.25 & 0.25 & 0.05  \\
batch size & 10 & 100 & 100 & 100 & 100 \\
iteration & 1000 & 1000 & 4000 & 4000 & 2000 \\
% \bottomrule
\end{NiceTabular}
}
% \vspace{-5pt}
\caption{Specific hyper-parameters employed in data synthesis.}
\label{tab:recover}
% \vspace{-.5em}
\end{table}

%% file: tables/tab-validation.tex
\begin{table}[h]
\centering
\resizebox{\linewidth}{!}{
\begin{NiceTabular}{c|ccccc}
% \toprule
config & CIFAR-10 & CIFAR-100 & Tiny-ImageNet & ImageNet-1K & ImageNet-21K \\
\midrule[1pt]
optimizer & \multicolumn{5}{c}{AdamW} \\
momentum & \multicolumn{5}{c}{$\beta_1$, $\beta_2$ = 0.9, 0.999}  \\
learning rate & \multicolumn{5}{c}{1e-3} \\
weight decay & \multicolumn{5}{c}{1e-2} \\
lr schedule & \multicolumn{5}{c}{cosine} \\
augmentation & \multicolumn{5}{c}{Random Resized Crop} \\
batch size & 16 & 64 & 64 & 128 & 32 \\
epoch & 1000 & 1000 & 500 & 300 & 300 \\
% \bottomrule
\end{NiceTabular}
}
% \vspace{-2pt}
\caption{Comprehensive hyper-parameter configuration for evaluation in downstream model training.}
\label{tab:validation}
% \vspace{-.5em}
\end{table}

%% file: figures/supp-compare.tex
\begin{figure}[!t]
\vspace*{-85pt} 
\begin{minipage}[t]{\linewidth}
    \centering
    \includegraphics[width=\linewidth]{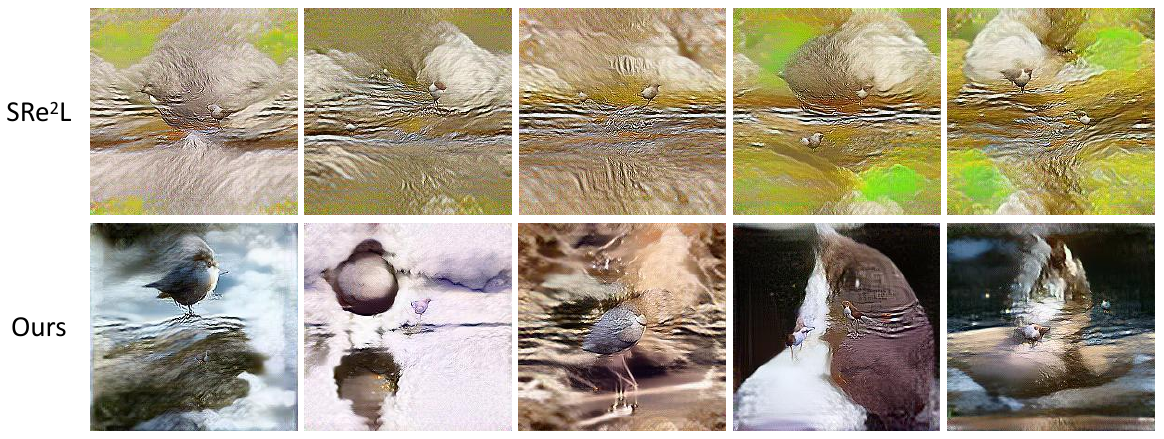}
    \centerline{(a) Water Ouzel}
    % \vspace{-2pt}
\end{minipage}\\
% \hfill
\begin{minipage}[t]{\linewidth}
    \centering
    \includegraphics[width=\linewidth]{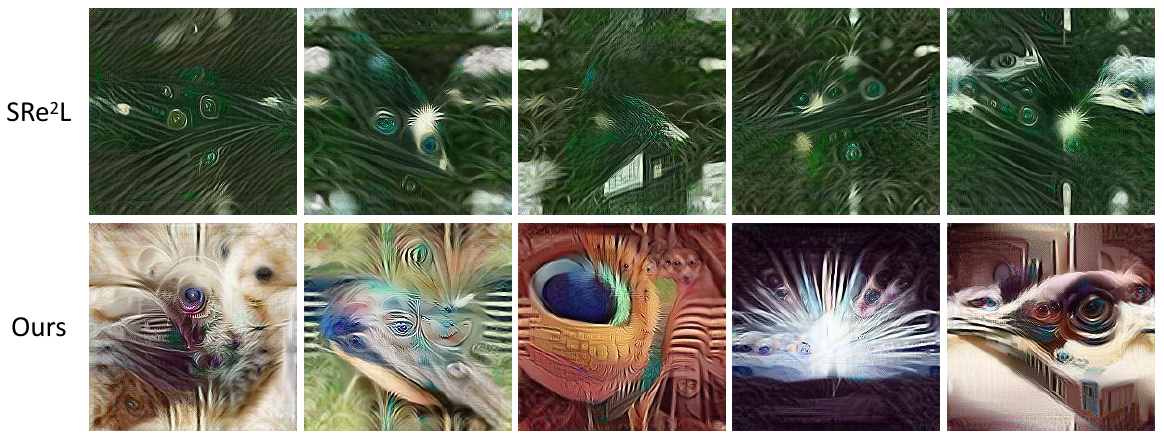}
    \centerline{(b) Peacock}
    % \vspace{-2pt}
\end{minipage}
% \hfill
\begin{minipage}[t]{\linewidth}
    \centering
    \includegraphics[width=\linewidth]{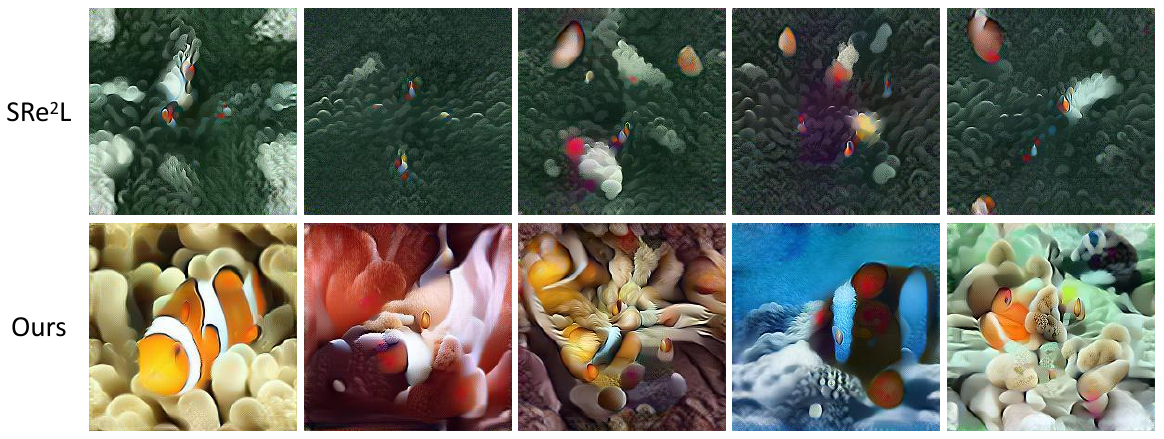}
    \centerline{(c) Anemone Fish}
    % \vspace{-2pt}
\end{minipage}
% \hfill
\begin{minipage}[t]{\linewidth}
    \centering
    \includegraphics[width=\linewidth]{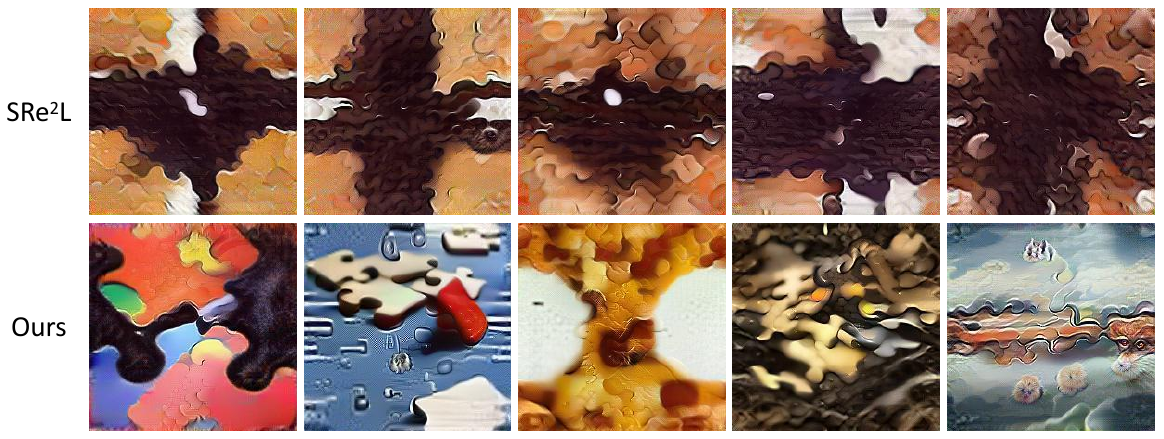}
    \centerline{(d) Jigsaw Puzzle}
    % \vspace{-2pt}
\end{minipage}
% \hfill
\begin{minipage}[t]{\linewidth}
    \centering
    \includegraphics[width=\linewidth]{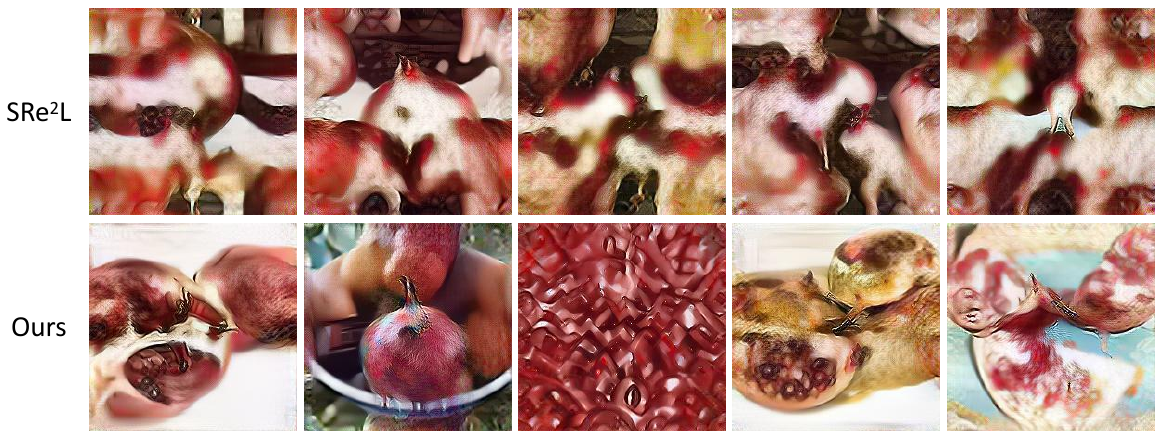}
    \centerline{(e) Pomegranate}
\end{minipage}
% \vspace{-5pt}
    \caption{Additional comparisons between synthetic data distilled by our method and SRe$^2$L.}
    \label{fig:supp-compare}
% \vspace{-.5em}
\end{figure}

%% file: figures/supp-vis-in1k.tex
\begin{figure*}[!t]
\centering
\includegraphics[width=0.95\linewidth]{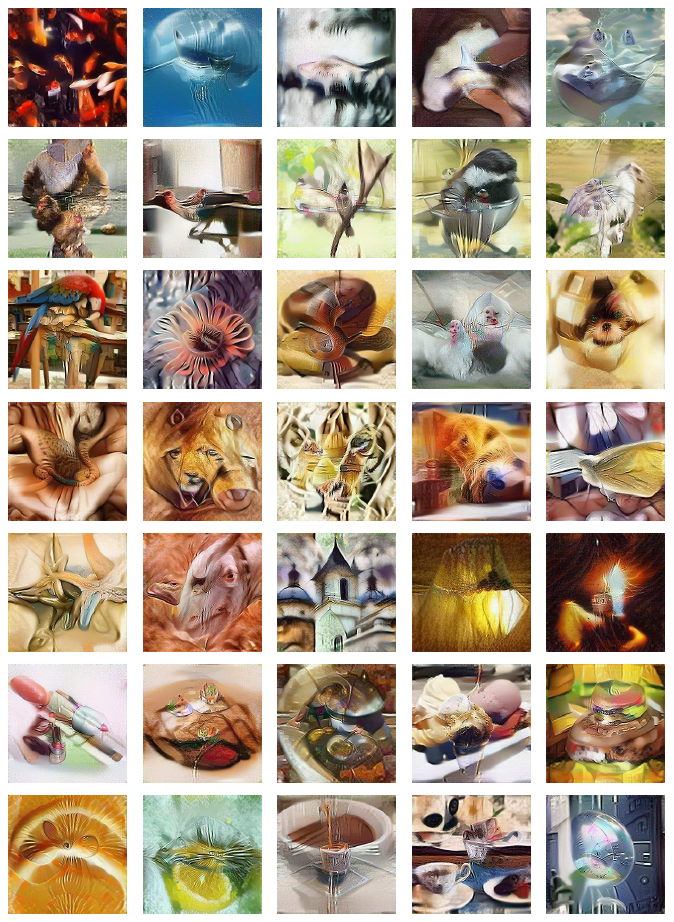}
% \vspace{4pt}
% \vspace{-8pt}
    \caption{Synthetic data on ImageNet-1K.}
    \label{fig:supp-vis-in1k}
% \vspace{-.5em}
\end{figure*}

%% file: figures/supp-vis-in21k.tex
\begin{figure*}[!t]
\centering
\includegraphics[width=0.95\linewidth]{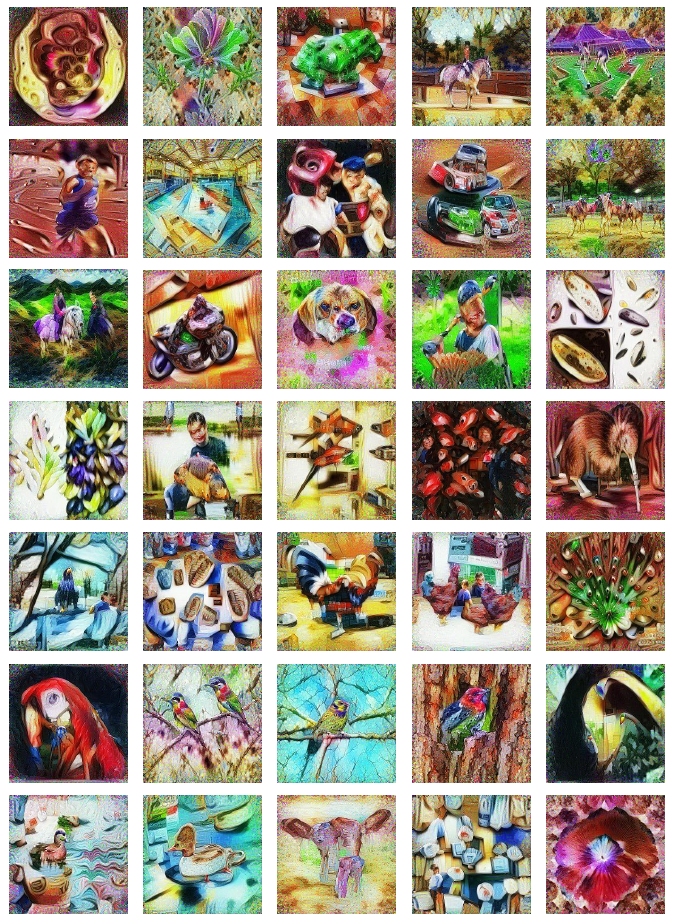}
% \vspace{4pt}
% \vspace{-8pt}
    \caption{Synthetic data on ImageNet-21K.}
    \label{fig:supp-vis-in21k}
% \vspace{-.5em}
\end{figure*}